\documentclass[12pt]{article}
\usepackage{amsmath}
\usepackage{amsthm}
\usepackage{amssymb}
\usepackage{graphicx}
\usepackage[margin=2cm]{geometry}
\usepackage{setspace}
\usepackage{epstopdf}
\usepackage{subfig}
\usepackage{algorithm}
\usepackage{algorithmicx}
\usepackage{algpseudocode}
\usepackage{multirow}
\usepackage{url}
\usepackage[title]{appendix}

\newcommand\MyHead[2]{%
  \multicolumn{1}{r}{\parbox{#1}{\centering #2}}
}

\makeatletter

\theoremstyle{plain}

  \theoremstyle{definition}
  
  \theoremstyle{plain}

\makeatother

  \providecommand{\definitionname}{Definition}
  \providecommand{\lemmaname}{Lemma}
\providecommand{\theoremname}{Theorem}

\renewcommand{\vec}[1]{{{#1}}}

\usepackage{endnotes}
\usepackage{color}
\usepackage{xparse}

\begin{document}

\title{Steerable Principal Components for Space-Frequency Localized Images}

\author{Boris Landa and Yoel Shkolnisky}


\maketitle

\bigskip

\noindent Boris Landa \\
Department of Applied Mathematics, School of Mathematical Sciences \\
Tel-Aviv University \\
{\tt sboris20@gmail.com}

\bigskip
\bigskip

\noindent Yoel Shkolnisky \\
Department of Applied Mathematics, School of Mathematical Sciences \\
Tel-Aviv University \\
{\tt yoelsh@post.tau.ac.il}

\bigskip
\bigskip

\begin{center}
Please address manuscript correspondence to Boris Landa,
{\tt sboris20@gmail.com}, (972)~549427603.
\end{center}
\newpage

\begin{abstract}
As modern scientific image datasets typically consist of a large number of images of high resolution, devising methods for their accurate and efficient processing is a central research task. In this paper, we consider the problem of obtaining the steerable principal components of a dataset, a procedure termed ``steerable-PCA''. The output of the procedure is the set of orthonormal basis functions which best approximate the images in the dataset and all of their planar rotations. To derive such basis functions, we first expand the images in an appropriate basis, for which the steerable-PCA reduces to the eigen-decomposition of a block-diagonal matrix. If we assume that the images are well localized in space and frequency, then such an appropriate basis is the Prolate Spheroidal Wave Functions (PSWFs). We derive a fast method for computing the PSWFs expansion coefficients from the images' equally-spaced samples, via a specialized quadrature integration scheme, and show that the number of required quadrature nodes is similar to the number of pixels in each image. We then establish that our PSWF-based steerable-PCA is both faster and more accurate then existing methods, and more importantly, provides us with rigorous error bounds on the entire procedure.
\end{abstract}

\section{Introduction}
Principal Component Analysis (PCA), also known as Karhunen-Loeve transform, is a ubiquitous method for dimensionality reduction which is often utilized for compression, de-noising, and feature-extraction from datasets. Given a dataset, the basic idea behind classical PCA is to find the best linear approximation (in the least-squares sense) to the dataset using a set of orthonormal basis functions, thus allowing for processing methods adaptive to the dataset at hand. Due to increasing improvements in image acquisition and storage techniques, we often encounter the need to process very large datasets, which consist of many thousands of images of ever-growing resolutions. In addition, there exist image acquisition techniques that introduce known deformation types into the images, thus increasing the variability in the data.

In this work, we focus on the rotation deformation, and specifically, on the setting where each image was acquired through an unknown planar rotation. Therefore, it is only natural to include all planar rotations of all images when performing the PCA procedure. It is important to mention that when handling large datasets, the naive approach of introducing a large number of rotated versions of all images into the dataset, and then performing standard PCA, is computationally prohibitive, and moreover, is less accurate than considering the continuum of all rotations.
In the literature, one can find numerous works concerning the study of deformations (such as rotations, translations, dilations, etc.) and their connections to invariant-feature extraction, through the theory of Lie groups and compact group theory~\cite{lenz1989group,ferraro1988relationship,lenz1990group,lenz1990group2,kanatani2012group}. In this context, we aim to incorporate the action of the group SO(2) into the framework of PCA of image datasets. We will refer to such a procedure as steerable-PCA. By the theory of Lie groups, and specifically the action of the group SO(2), it is known that the resulting principal components (which best approximate all rotations of a set of images) are described by tensor products of radial functions and angular Fourier modes. Such functions are referred to as ``steerable''~\cite{freeman1991design}, since they can be rotated (or ``steered'') by a simple multiplication with a complex-valued constant, and hence the term ``steerable'' in steerable-PCA. An early approach for computing the steerable-PCA was proposed in~\cite{perona1995deformable}, which used the SVD to obtain the principal components of a single template and its set of deformations.
In~\cite{hilai1994recognition}, the authors presented an efficient steerable-PCA algorithm based on angular Fourier decomposition of the images, after they have been resampled to a polar grid. While this method considers the continuum of all planar rotations, it requires a non-obvious discretization of the images in the radial direction. Efficient methods for steerable-PCA were also introduced in~\cite{ponce2011computing} and~\cite{jogan2002karhunen}. These methods considered a finite set of equiangular rotations of each image on a polar grid, which allows for efficient circulant matrix decompositions to be carried out when computing the principal components. Recently,~\cite{zhao2013fourier} and~\cite{zhao2014fast} utilized Fourier-Bessel basis functions to expand the images, followed by applying PCA in the domain of the expansion coefficients, thus accounting for all (infinitely many) rotations. We also mention~\cite{vonesch2015steerable}, where the authors present an accurate algorithm to obtain the steerable principal components of templates whose analytical form is known in advance.

As digital images are typically specified by their samples on a Cartesian grid, considering their rotations implicitly assumes that they were sampled from some underlying bivariate function. Rotation of a sampled image essentially interpolates this underlying function from the available Cartesian samples.
We remark that while previous works provide algorithms for steerable-PCA of discretized input images, they lack in describing rigorous connections between the results of the procedure and the images  prior to discretization, i.e. the underlying bivariate functions.
In this work, we assume that the digital images were sampled from bivariate functions that are essentially bandlimited and are sufficiently concentrated in an area of interest in space. These assumptions guarantee that if an image was sampled with a sufficiently high sampling rate, it can be reconstructed from its samples with high precision~\cite{petersen1962sampling}.
Such assumptions are very common in various areas of engineering and physics, and as all acquisition devices are essentially bandlimited and restricted in space/time, are expected to hold for a wide range of image datasets.
Since we are interested in processing images with arbitrary orientations, it is only natural to consider a circular support area, both in space and frequency, instead of the (classical) square.
We note that this model was implicitly assumed to hold in~\cite{zhao2013fourier} and~\cite{zhao2014fast} for single-particle cryo-electron microscopy (cryo-EM) images.

Under the model assumptions mentioned above, our goal is to develop a fast and accurate steerable-PCA procedure which considers the continuum of all planar rotations of all images in our dataset.
Particularly appealing basis functions for expanding bandlimited functions which are also concentrated in space, are the Prolate Spheroidal Wave Functions (PSWFs)~\cite{slepian1961prolate,landau1961prolate,landau1962prolate,slepian1964prolate,osipov2013prolate}, defined as the strictly bandlimited set of orthogonal functions, which maximize the ratio between their $\mathcal{L}^2$ norm inside some finite region of interest and their $\mathcal{L}^2$ norm over the entire Euclidean space. Recently, \cite{landa2016approximation} described an approximation scheme for functions localized to disks in space and frequency, using a series of two-dimensional PSWFs. We therefore incorporate the methods introduced in~\cite{landa2016approximation} into the framework of steerable-PCA, providing accurate, scalable and efficient algorithms. Our approach is in the spirit of~\cite{zhao2014fast}, resulting in a similar block-diagonal covariance matrix. However, replacing the Fourier-Bessel with PSWFs turns out to be advantageous in terms of accuracy, available error bounds, speed, and statistical properties.

The contributions of this paper are the following. By utilizing theoretical and computational tools related to PSWFs, we are able to provide accuracy guarantees for our steerable-PCA algorithm, under the assumptions of space-frequency localization. This accuracy is in part related to a rigorous truncation rule we provide for the PSWFs series expansion, in contrast to the series truncation rules used in~\cite{zhao2013fourier} and~\cite{zhao2014fast}. In addition, using a quadrature integration scheme optimized for integrating bandlimited functions on a disk~\cite{shkolnisky2007prolate}, we present an algorithm which is in theory (and in practice) faster than~\cite{zhao2014fast} by a factor between $2$ and $4$. Finally, we also show that under some conditions on the space-frequency concentration of the images at hand, the transformation to the PSWFs expansion coefficients is nearly orthogonal.

The organization of this paper is as follows. In Section~\ref{section:Image approximation based on PSWFs expansion} we introduce the PSWFs and their usage in expanding space-frequency localized images specified by their equally-spaced Cartesian samples. In particular, we review the results of~\cite{landa2016approximation} on expanding a function into a series of PSWFs, evaluating the expansion coefficients, and bounding the overall approximation error. In Section~\ref{section:Fast PSWFs coefficients approximation} we present a fast algorithm for approximating the expansion coefficients from Section~\ref{section:Image approximation based on PSWFs expansion} up to an arbitrary precision. In Section~\ref{section:Steerable PCA procedure} we formalize the procedure of steerable-PCA for the continuous setting (similarly to~\cite{vonesch2015steerable}), and combine it with our PSWFs-based approximation scheme. Section~\ref{section:algorithm summary and computational cost} then summarizes all relevant algorithms, and analyses in detail the computational complexities involved. In Section~\ref{section:transformation spectrum} we provide some numerical results on the spectrum of the transformation to the PSWFs expansion coefficients, as this spectrum is of particular interest for noisy datasets, and in Section~\ref{section:Numerical experiments} we compare our algorithm to that of~\cite{zhao2014fast} in terms of running time and accuracy. Finally, in Section~\ref{section:Summary and discussion} we provide some concluding remarks and some possible future research directions.

\section{Image approximation based on PSWFs expansion} \label{section:Image approximation based on PSWFs expansion}
\begin{figure}
  \centering
  	\subfloat[$\psi_{N,n}^c(x), x\in\mathbb{R}^2$]{
    \includegraphics[width=0.5\textwidth]{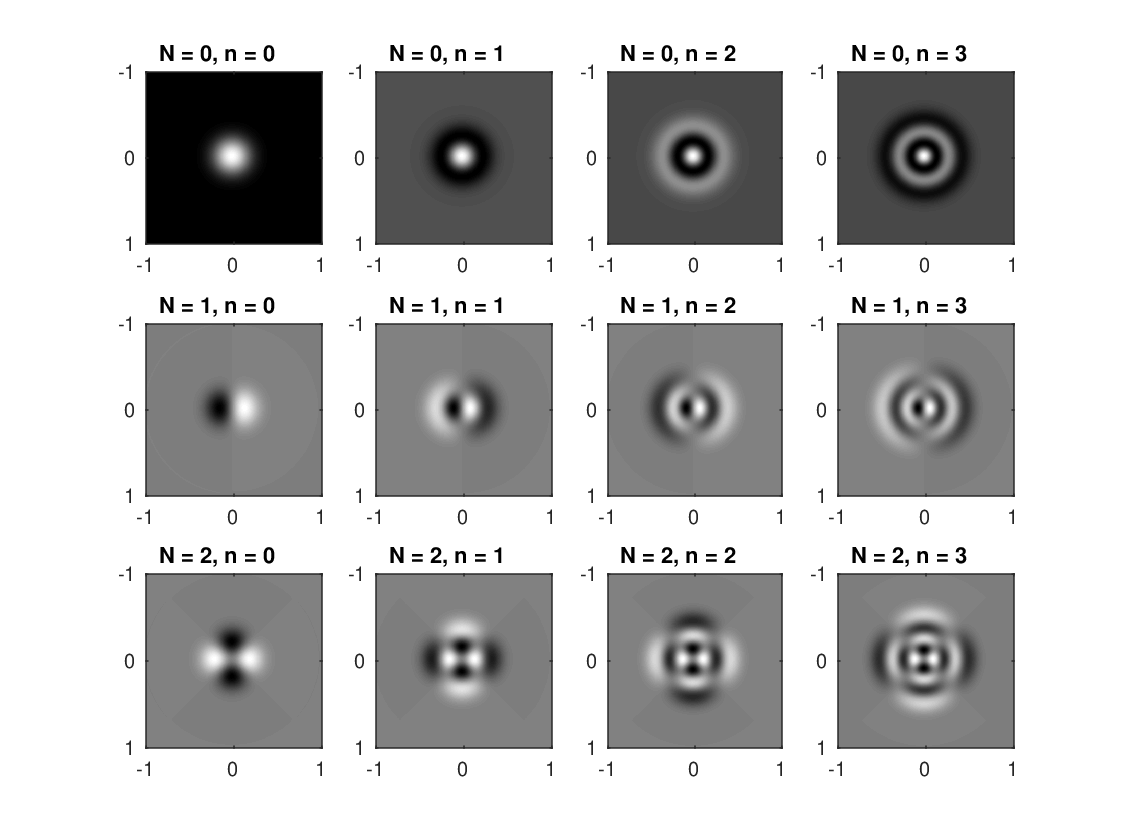}
    }
    \subfloat[$R_{N,n}^c(r)$]{ \label{fig:pswfs illustration radial profile}
    \includegraphics[width=0.5\textwidth]{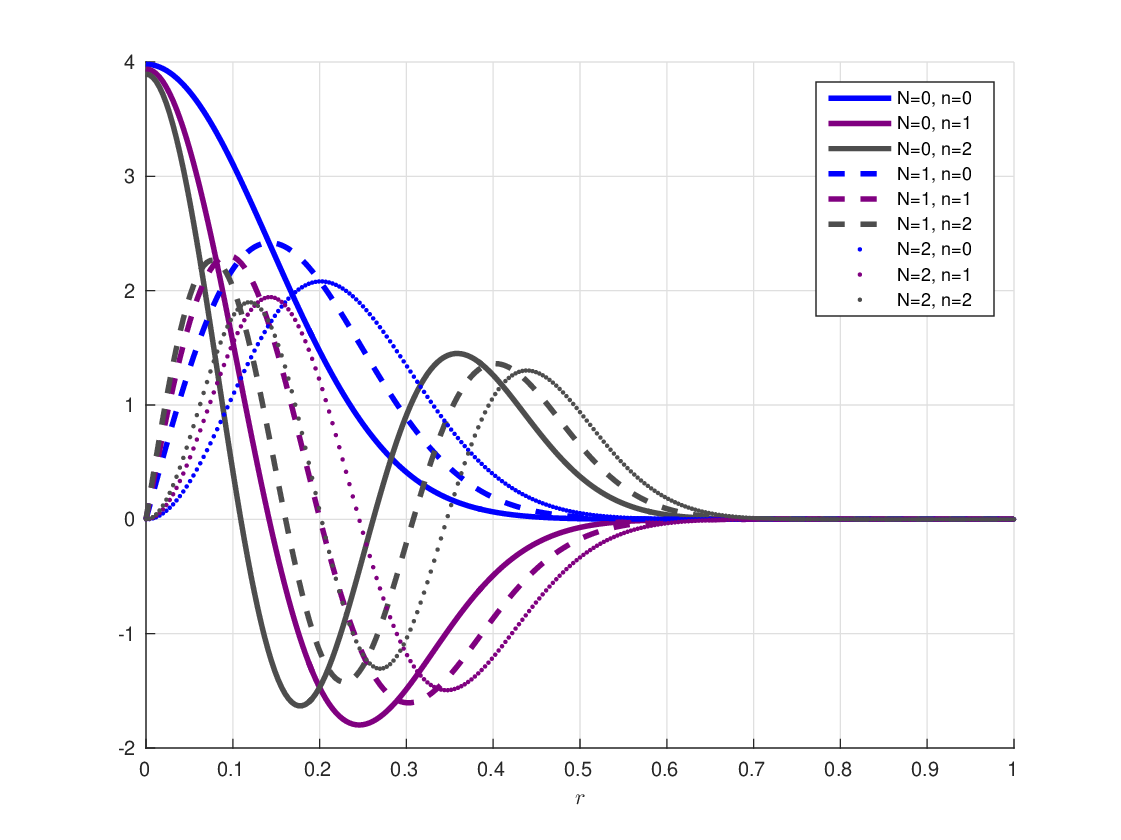}
    }
	\caption[pswf_radial_profile]
	{Illustration of the first few PSWFs (real part) with bandlimit $c=16\pi$, ordered according to their eigenvalues (for every index $N$, the eigenvalues are ordered from largest to smallest as a function of the index $n$).} \label{fig:pswfs illustration}
\end{figure}

Let $f:\mathbb{R}^2 \rightarrow \mathbb{R}$ be a square integrable function on $\mathbb{R}^2$.
We define a function $f(x)$ as $c$-bandlimited if its two-dimensional Fourier transform, denoted by $F(\omega)$, vanishes outside a disk of radius $c$. Specifically, if we denote $\mathbf{D}\triangleq\{x\in\mathbb{R}^{2},\:\left|x\right|\leq1\}$, then $f$ is bandlimited with bandlimit $c$ if
\begin{equation}
f(x)= \left(\frac{1}{2\pi}\right)^{2} \int_{c\mathbf{D}} F(\omega)e^{\imath \omega x}d\omega, \quad x \in\mathbb{R}^{2}.\label{eq:Bandlimited function}
\end{equation}
Among all $c$-bandlimited functions, the Prolate Spheroidal Wave Functions (PSWFs) on $\mathbf{D}$ (the unit disk), are the most energy concentrated in $\mathbf{D}$, that is, they satisfy
\begin{equation}
\frac{\left\Vert \psi (x)\right\Vert_{\mathcal{L}^2(\mathbf{D})}}{\left\Vert \psi (x)\right\Vert_{\mathcal{L}^2(\mathbb{R}^2)}} \rightarrow \underset{\psi}{\max},
\end{equation}
while constituting an orthonormal system over ${\mathcal{L}^2(\mathbf{D})}$. The two-dimensional PSWFs were derived and analysed in \cite{slepian1964prolate}, and were shown to be the solutions to the integral equation
\begin{equation}
\alpha\psi(x)=\int_{\mathbf{D}}\psi(\omega)e^{\imath c \omega x}d\omega ,\quad x\in\mathbf{D}.\label{eq:Basic PSWF eq}
\end{equation}
We denote the PSWFs with bandlimit $c$ as $\psi_{N,n}^c(x)$, and their corresponding eigenvalues as $\alpha_{N,n}^c$,
which together form the eigenfunctions and eigenvalues of~\eqref{eq:Basic PSWF eq}, with $N\in\mathbb{Z}$ and $n\in\mathbb{N}\cup\left\lbrace 0 \right\rbrace$. In addition, it turns out that the functions $\psi_{N,n}^c(x)$ are orthogonal on
both $\mathbf{D}$ and $\mathbb{R}^{2}$ using the standard $\mathcal{L}^2$ inner products on $\mathbf{D}$ and $\mathbb{R}^{2}$, respectively, and are dense in both the
class of $\mathcal{L}^{2}(\mathbf{D)}$ functions and in the class of $c$-bandlimited
functions on $\mathbb{R}^{2}$.
In polar coordinates, the functions $\psi_{N,n}^c(x)$ have a separation of variables and can be written in complex form as
\begin{equation}
\psi_{N,n}^c(r,\theta)=\frac{1}{\sqrt{2\pi}}R_{N,n}^c(r)e^{\imath N\theta}, \quad N\in\mathbb{Z}, \ n\in\mathbb{N}\cup\left\lbrace 0 \right\rbrace, \label{eq:PSWFs complex form}
\end{equation}
where $R_{N,n}(r)$ (defined explicitly in~\eqref{eq:PSWFs radial part integral eq} in Appendix~\ref{appendix:Behevaior of lambda}) satisfies $R_{N,n}(r) = R_{\left|N\right|,n}(r)$, and the eigenfunctions $\psi_{N,n}^c(x)$ are normalized to have an $\mathcal{L}^2(\mathbf{D})$ norm of $1$. The indices $N$ and $n$ are often referred to as the angular index and the radial index respectively. Equation~\eqref{eq:PSWFs complex form} also tells us that the PSWFs are steerable~\cite{freeman1991design}, that is, rotating $\psi_{N,n}^c(r,\theta)$ is equivalent to multiplying it by a complex constant. This property is important for handling datasets which include rotations, and in particular, for the steerable-PCA procedure.
A detailed numerical evaluation procedure for the two-dimensional PSWFs can be found in \cite{shkolnisky2007prolate}, and an illustration of the PSWFs for the several first index pairs $(N,n)$ can be seen in Figure~\ref{fig:pswfs illustration}.

Since our  images are assumed to be essentially bandlimited and sufficiently concentrated in a disk, the properties of the PSWFs mentioned above (and especially their optimal concentration property) make them suitable basis functions for expanding such images, as we consider next.
Let us define a function $f(x)$ as $\left( \nu,\mu \right)$-concentrated if its $\mathcal{L}^2$ norm outside a disk of radius $\nu$ is upper bounded by $\mu$, that is
\begin{equation}
\sqrt{\int_{x \notin \nu \mathbf{D}} {\left| f(x) \right|}^2 dx} \leq \mu.
\end{equation}
Using this definition, the class of $c$-bandlimited functions is a subclass of $\left( c,\delta_c \right)$-concentrated functions in the Fourier domain (for any $\delta_{c}\geq 0$).
We consider the two-dimensional functions from which our images were sampled to be $\left( 1,\varepsilon \right)$-concentrated in space, with their Fourier transforms being $\left( c,\delta_c \right)$-concentrated. This assumption always holds for some set of parameters $c$, $\delta_c$, $\varepsilon$, and the general notion is that $\delta_c$ and $\varepsilon$ are "small".
Since our images are given in their sampled form, we define the unit square $\mathbf{Q} \triangleq \left[-1,1\right]\times\left[-1,1\right]$, and assume that we are given the samples $\left\lbrace f(x_\vec{k}): x_\vec{k}=\frac{\vec{k}}{L} \in \mathbf{Q} \right\rbrace$, where $\vec{k}$ is a two-dimensional integer index. These samples correspond to a Cartesian grid of $\left(2L+1\right)\times\left(2L+1\right)$ equally spaced samples, with sampling frequency of $L$ in each dimension.
We mention that in this setup, the Nyquist frequency corresponds to a bandlimit $c$ of at most $\pi L$.

\begin{figure}
  \centering  	
    \includegraphics[width=0.6\textwidth]{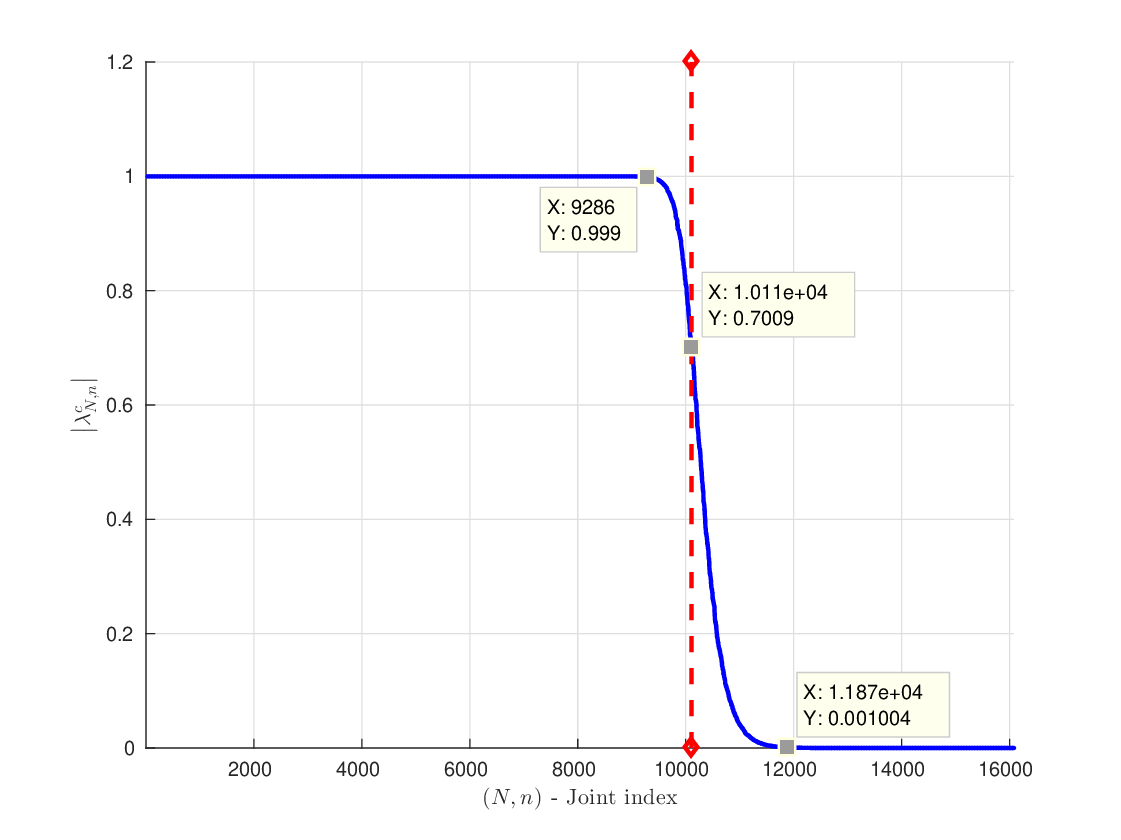}
	\caption[pswf_lambda_Nn]
	{Illustration of the normalized eigenvalues $\left| \lambda_{N,n}^c \right|$ for $L=64$, $c=\pi L$, sorted in a non-increasing order with a joint index $k$ enumerating over $(N,n)$. It is evident that for $T \ll 1$ the set $\Omega_T$ consists of indices up to the right of the vertical dashed line, where the normalized eigenvalues rapidly decay to zero. On the other hand, for $T \gg 1$ the set $\Omega_T$ consists of indices up to the left of the vertical line, where the normalized eigenvalues are extremely close to one. The dashed vertical line corresponds to $k={c^2}/{4}$, which also approximately agrees with $\left| \lambda_{N,n}^c \right|^2 = 0.5$ (or equivalently with the rule $T=1$).} \label{fig:pswfs lambda_Nn behavior}
\end{figure}

Given an image $I(x)\in\mathcal{L}^2(\mathbf{D})$, we can expand it as
\begin{equation} \label{eq: Infinite PSWF expansion}
I(x) = \sum_{N= -\infty}^{\infty} \sum_{n=0}^{\infty} a_{N,n} \psi_{N,n}^c(x), \quad x \in \mathbf{D}, \quad \qquad a_{N,n} = \int_{\mathcal{\mathbf{D}}} I(x) \left(\psi_{N,n}^c(x)\right)^* dx,
\end{equation}
where $(\cdot)^*$ denotes complex conjugation.
As numerical algorithms cannot use infinite expansions, the image $I(x)$ needs to be approximated by a finite sequence of PSWFs, while its expansion coefficients $\left\lbrace a_{N,n} \right\rbrace$ are approximated using only the available samples $\left\lbrace I(\frac{k}{L}) \right\rbrace_{\frac{k}{L} \in \mathbf{Q}}$.
To this end, we follow~\cite{landa2016approximation}, and define the truncated PSWFs expansion that approximates $I(x)$ as
\begin{equation} \label{eq:Coeff approx series}
\hat{I}(x) \triangleq \underset{N,n \in \Omega_T}{\sum}\hat{a}_{N,n}{\psi}_{N,n}^c(x),
\end{equation}
where $\hat{a}_{N,n}$ are the approximated coefficients (to be described shortly), and $\Omega_T$ is a finite set of indices that is determined by a truncation parameter $T$, and is defined by
\begin{equation} \label{eq:PSWFs truncation rule}
\Omega_{T}\triangleq\left\{ (N,n):\;\sqrt{\frac{\left\vert\lambda_{N,n}^{c}\right\vert^2}{1-\left\vert\lambda_{N,n}^{c}\right\vert^2}}>T\right\}, \quad T>0 , \qquad \qquad \lambda_{N,n}^c \triangleq \frac{c}{2\pi} { \alpha_{N,n}^c },
\end{equation}
where $\alpha_{N,n}^c$ is the eigenvalue from~\eqref{eq:Basic PSWF eq} corresponding to the eigenfunction $\psi_{N,n}^c$. The properties of the index set $\Omega_T$ are determined by the ''normalized'' eigenvalues $\lambda_{N,n}^c$, whose behaviour is exemplified in Figure~\ref{fig:pswfs lambda_Nn behavior}.
The coefficients $\hat{a}_{N,n}$ of~\eqref{eq:Coeff approx series} are defined by
\begin{equation}
\hat{a}_{N,n} \triangleq  \left\vert\lambda_{N,n}^{c}\right\vert^2 \frac{\left\langle I,{\psi}_{N,n}^c \right\rangle}{L^2}   =  \frac{\left\vert\lambda_{N,n}^{c}\right\vert^2}{L^2}  \underset{\frac{k}{L}\in\mathbf{D}}{\sum}I(\frac{k}{L})\left( {\psi}^c_{N,n}(\frac{k}{L})\right)^*, \label{eq:Modified Prolates coefficient evaluation}
\end{equation}
where both $I$ and ${\psi}_{N,n}^c$ stand for the appropriate vectors of samples of $I(x)$ and ${\psi}_{N,n}^c(x)$ inside the unit disk, respectively. We note that for real-valued images we have that $a_{N,n}=\left(a_{-N,n}\right)^*$, and thus it is sufficient to compute the coefficients $\hat{a}_{N,n}$ only for $N\geq 0$.

 Using our definitions in \eqref{eq:Coeff approx series}, \eqref{eq:PSWFs truncation rule}, \eqref{eq:Modified Prolates coefficient evaluation}, and assuming that $I(x)$ is $(1,\varepsilon)$-concentrated in space and $(c,\delta_c)$-concentrated in Fourier domain (as defined earlier), we show in Appendix~\ref{appendix:PSWFs expansion error bound} that
\begin{equation}
{\left\Vert I(x) - \underset{N,n \in \Omega_T}{\sum}\hat{a}_{N,n}{\psi}_{N,n}^c(x) \right\Vert}_{\mathcal{L}^2\left( \mathbf{D}\right)}
\leq
\left( \varepsilon + \frac{\delta_c}{2\pi} \right) \left( T + 4\right) \; \triangleq \; \mathcal{E}(\varepsilon,\delta_c,T). \label{eq:Total approx err 2}
\end{equation}
It is evident that the bound $\mathcal{E}(\varepsilon,\delta_c,T)$ in~\eqref{eq:Total approx err 2} depends only on the images' concentration properties and on the truncation parameter $T$, all of which can be determined a priori.

Lastly, it is also mentioned in~\cite{landa2016approximation} that the cardinality of the index set $\Omega_T$ of~\eqref{eq:PSWFs truncation rule} is expected to be
\begin{equation}
\left\vert \Omega_T \right| = \frac{c^2}{4} - \frac{2}{\pi^2} c\log{(c)}\log{(T)} + o(c\log{(c)}). \label{eq:Asymptotic number of terms}
\end{equation}
The first term on the right hand-side of~\eqref{eq:Asymptotic number of terms}, which depends quadratically on $c$, is in-fact (up to a constant of ${1}/{\pi}$) what is known as the Landau rate~\cite{landau1967necessary} for stable sampling and reconstruction, and it is the term which dominates asymptotically the required number of basis functions for the approximation (see also Figure~\ref{fig:pswfs lambda_Nn behavior}). A table listing the values of $\left\vert \Omega_T \right|$ for various $L$ and $T$ can be found in~\cite{landa2016approximation}.
We remark that for $c=\pi L$ (Nyquist sampling) and $T=1$, which essentially means that the approximation error is of the order of the space-frequency localization, we have that $\left\vert \Omega_T \right|$ is about ${\pi^2}L^2/4$, which is smaller than the number of samples (pixels) in the unit disk.

\section{Fast PSWFs coefficients approximation} \label{section:Fast PSWFs coefficients approximation}
Computing the expansion coefficients $\hat{a}_{N,n}$ of~\eqref{eq:Modified Prolates coefficient evaluation} directly, results in a computational complexity of $O(L^4)$ operations. This is due to the fact that each image contains $O(L^2)$ samples, while there are about $O(L^2)$ different basis functions in the expansion. Since we aim to process large datasets of high resolution images, in what follows, we describe an asymptotically more efficient method for evaluating the expansion coefficients in $O(L^3)$ operations.

\begin{figure}
  \centering  	
    \includegraphics[width=0.5\textwidth]{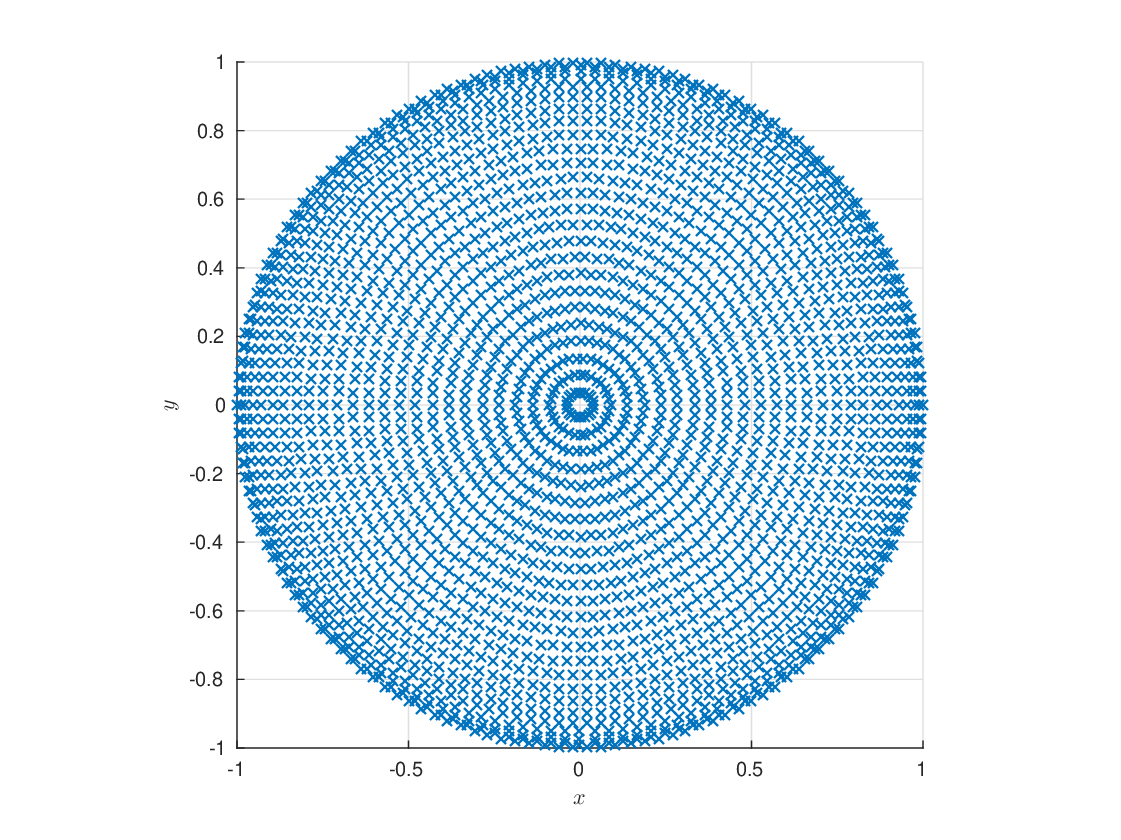}
	\caption[quadrature_nodes]
	{Quadrature nodes $\omega_{\ell,j}^{2c} = \left( r_\omega^\ell, \theta_\omega^{\ell,j} \right)$ for $L=16$, $c=\pi L$, and accuracy of the order of machine precision. The nodes are arranged to be equally-spaced in the angular direction (the specific phase of the allocation is arbitrary) with different numbers of nodes for each radius (rings closer to the origin require less angular nodes). The number of radial nodes is about ${c}/{\pi}$, and the number of angular nodes per radius $r_\omega^\ell$ is about $c r_\omega^\ell e$, where $e$ is the base of the natural logarithm.} \label{fig:quadrature nodes}
\end{figure}

Using~\eqref{eq:Basic PSWF eq}, we can rewrite the PSWFs expansion coefficients~\eqref{eq:Modified Prolates coefficient evaluation} as
\begin{align}
\hat{a}_{N,n} &= \alpha_{N,n}^c\left(\frac{c}{2\pi L}\right)^2  \underset{\frac{k}{L}\in\mathbf{D}}{\sum}I(\frac{k}{L})\left[ \int_\mathcal{\mathbf{D}} \psi_{N,n}^c(\frac{k}{L}) e^{\imath c \omega \frac{k}{L}} \right]^* \\
&= \alpha_{N,n}^c\left(\frac{c}{2\pi L}\right)^2 \int_{\mathbf{D}}\left[ \psi^c_{N,n}(\omega) \right]^*  \phi^c(\omega) d\omega, \label{eq:Coefficients integral evaluation}
\end{align}
where
\begin{equation}
\phi^c(u) \triangleq \underset{\frac{k}{L}\in\mathbf{D}}{\sum}I(\frac{k}{L}) e^{-\imath c u \frac{k}{L}}, \quad u\in\mathbb{R}^2.  \label{eq:Phi def}
\end{equation}
Since both $\psi^c_{N,n}$ and $\phi^c$ are $c$-bandlimited functions, the product $\psi^c_{N,n}\cdot \phi^c$ defines a function which is $2c$-bandlimited. Therefore, we can compute the integral in~\eqref{eq:Coefficients integral evaluation} using a quadrature formula for $2c$-bandlimited functions on a disk. Such an integration scheme was proposed in \cite{shkolnisky2007prolate} using specialized quadrature nodes and weights. Thus, it follows that by using such a quadrature formula, we can approximate the expansion coefficients of~\eqref{eq:Coefficients integral evaluation} by
\begin{equation}
\tilde{a}_{N,n} \triangleq \alpha_{N,n}^c\left(\frac{c}{2\pi L}\right)^2 \sum_{\ell = 1}^{\mathcal{N}_r} \sum_{j = 1}^{\mathcal{N}_\theta^{\ell}} \mathcal{W}_{\ell,j}^{2c} \left[ \psi^c_{N,n}(\omega_{\ell,j}^{2c}) \right]^*  \phi^c(\omega_{\ell,j}^{2c}), \label{eq:Bandlimited numerical integration formula}
\end{equation}
where $\omega_{\ell,j}^{2c}$ and $\mathcal{W}_{\ell,j}^{2c}$ are the quadrature nodes and weights respectively, $\mathcal{N}_r$ is the number of different radii of the quadrature nodes, and $\mathcal{N}_\theta^{\ell}$ is the number of quadrature nodes per radius (which may vary for each radius). We use for $\omega_{\ell,j}^{2c}$ and $\mathcal{W}_{\ell,j}^{2c}$ the quadrature nodes and weights of~\cite{shkolnisky2007prolate} corresponding to bandlimit $2c$. These quadrature nodes are equally-spaced in the angular direction and non-equally spaced in the radial direction. In polar coordinates, the nodes and weights are given by
\begin{equation}
\omega_{\ell,j}^{2c} = \left( r_\omega^\ell, \theta_\omega^{\ell,j} \right), \quad \quad \theta_\omega^{\ell,j} = \frac{2\pi j}{\mathcal{N}_\theta^{\ell}}, \quad \quad \mathcal{W}_{\ell,j}^{2c} = \frac{2\pi}{\mathcal{N}_\theta^{\ell}} r_\omega^\ell \tilde{\mathcal{W}}_{\ell}^{2c}, \label{eq:quadrature nodes polar coordinates}
\end{equation}
and the derivation of $\tilde{\mathcal{W}}_{\ell}^{2c}$ is detailed in \cite{shkolnisky2007prolate}. A typical array of quadrature nodes is plotted in Figure~\ref{fig:quadrature nodes}.

By employing~\eqref{eq:Bandlimited numerical integration formula},~\eqref{eq:quadrature nodes polar coordinates}, and the analytical expression of the PSWFs \eqref{eq:PSWFs complex form}, we finally arrive at
\begin{align}
\tilde{a}_{N,n} =
\sqrt{2\pi} \alpha_{N,n}^c \left(\frac{c}{2\pi L}\right)^2 \sum_{\ell = 1}^{\mathcal{N}_r} \tilde{\mathcal{W}}_{\ell}^{2c} \frac{R_{N,n}^c(r_\omega^\ell) r_\omega^\ell}{\mathcal{N}_\theta^{\ell}} \sum_{j = 1}^{\mathcal{N}_\theta^{\ell}} \phi^c(\omega_{\ell,j}^{2c}) e^{- \frac{2\pi \imath j N}{\mathcal{N}_\theta^{\ell}} }, \label{eq:expansion coeffs quadrature scheme}
\end{align}
which is particularly appealing for efficient numerical evaluation, as we explain later in this section.

Clearly, replacing the integral in~\eqref{eq:Coefficients integral evaluation} with a quadrature formula results in an approximation error. In this respect, it is desirable to choose $\mathcal{N}_r$ and $\mathcal{N}_\theta^{\ell}$ such that this error is smaller than some prescribed accuracy $\vartheta_q$
(usually chosen as the machine precision).
To this end, and according to \cite{shkolnisky2007prolate}, it is sufficient to satisfy the condition
\begin{equation}
\sum_{|j|>\mathcal{N}_\theta^{\ell}} J_j(2c r_\omega^\ell \rho) \leq C_1 \vartheta_q, \label{eq:numerical integration angular nodes condition}
\end{equation}
for all $1 \leq \ell \leq \mathcal{N}_r$ and $\rho \in [0,1]$, where $J_j$ is the Bessel function of the first kind and order $j$, as well as the condition
\begin{equation}
\sum_{k=2\mathcal{N}_r+1}^{\infty}  \frac{\left| \lambda_{0,k}^{2c} \right|}{2c}  \left\Vert R_{0,k}^{2c}(r) \right\Vert_{\infty} \left\Vert R_{0,k}^{2c}(r)\sqrt{r} \right\Vert_{\infty} \left( 1 + \sum_{j=1}^{\mathcal{N}_r} \left| \tilde{\mathcal{W}}_{j}^{2c} \right| \right) \leq C_2 \vartheta_q, \label{eq:numerical integration radial nodes condition}
\end{equation}
where $\lambda_{N,n}^c$ was defined in \eqref{eq:PSWFs truncation rule}, $\left\Vert \cdot \right\Vert_\infty$ stands for the max-norm in $[0,1]$, and both $C_1$ and $C_2$ are constants which depend on the bandlimit $c$. We mention that $\left| \lambda_{0,k}^{2c} \right|$ are ordered in a non-increasing order with respect to the index $k$. In order to determine appropriate values for $\mathcal{N}_r$ and $\mathcal{N}_\theta^{\ell}$, one can proceed to solve the inequalities in conditions~\eqref{eq:numerical integration angular nodes condition} and~\eqref{eq:numerical integration radial nodes condition} numerically by directly evaluating these expressions.
However, in order to analyse the computational complexity of the procedure described in this section, and to offer the reader simpler analytic expressions and some insight concerning conditions \eqref{eq:numerical integration angular nodes condition} and \eqref{eq:numerical integration radial nodes condition}, we provide some further analysis of these conditions and the resulting number of quadrature nodes. In Appendix~\ref{appendix:Number of quadrature nodes in theta}, we prove that in order to satisfy condition~\eqref{eq:numerical integration angular nodes condition} for every $1\leq\ell\leq\mathcal{N}_r$, it is sufficient to choose the integers $\mathcal{N}_\theta^{\ell}$ as
\begin{equation}
\mathcal{N}_\theta^{\ell} = \lceil {c r_\omega^\ell e} + \log{\vartheta_q^{-1}} + \log{\frac{2}{C_1}} + 1\rceil, \label{eq:numerical integration angular nodes analytic condition}
\end{equation}
where $r_\omega^\ell$ are the radial quadrature nodes from \eqref{eq:quadrature nodes polar coordinates}, $e$ is the base of the natural logarithm, and $\lceil \cdot \rceil$ is the rounding up operation. Then, it is evident that the term $c r_\omega^\ell e$ dominates condition \eqref{eq:numerical integration angular nodes analytic condition}, i.e. $\mathcal{N}_\theta^{\ell} \sim c r_\omega^\ell e$, since all other factors (prescribed error $\vartheta_q$ and the constant $C_1$) affect it only logarithmically.
Therefore, we expect the overall number of quadrature nodes to be
\begin{equation}
\sum_{\ell=1}^{\mathcal{N}_r} \mathcal{\mathcal{N}_\theta^{\ell}} \sim {ce\sum_{\ell=1}^{\mathcal{N}_r}  r_\omega^\ell \sim \frac{ce\mathcal{N}_r}{2}}, \label{eq:number of quad nodes 1}
\end{equation}
assuming that the quadrature points $r_\omega^\ell$ are approximately symmetric about $0.5$.
We remark that numerical experiments reveal that there is a small difference between conditions~\eqref{eq:numerical integration angular nodes condition} and~\eqref{eq:numerical integration angular nodes analytic condition}, and specifically, choosing $\mathcal{N}_\theta^{\ell}$ directly via the numerical evaluation of condition~\eqref{eq:numerical integration angular nodes condition} results in about $20\%$ less angular nodes compared to choosing $\mathcal{N}_\theta^{\ell}$ according to~\eqref{eq:numerical integration angular nodes analytic condition}.

\begin{figure}
  \centering
    \includegraphics[width=1\textwidth]{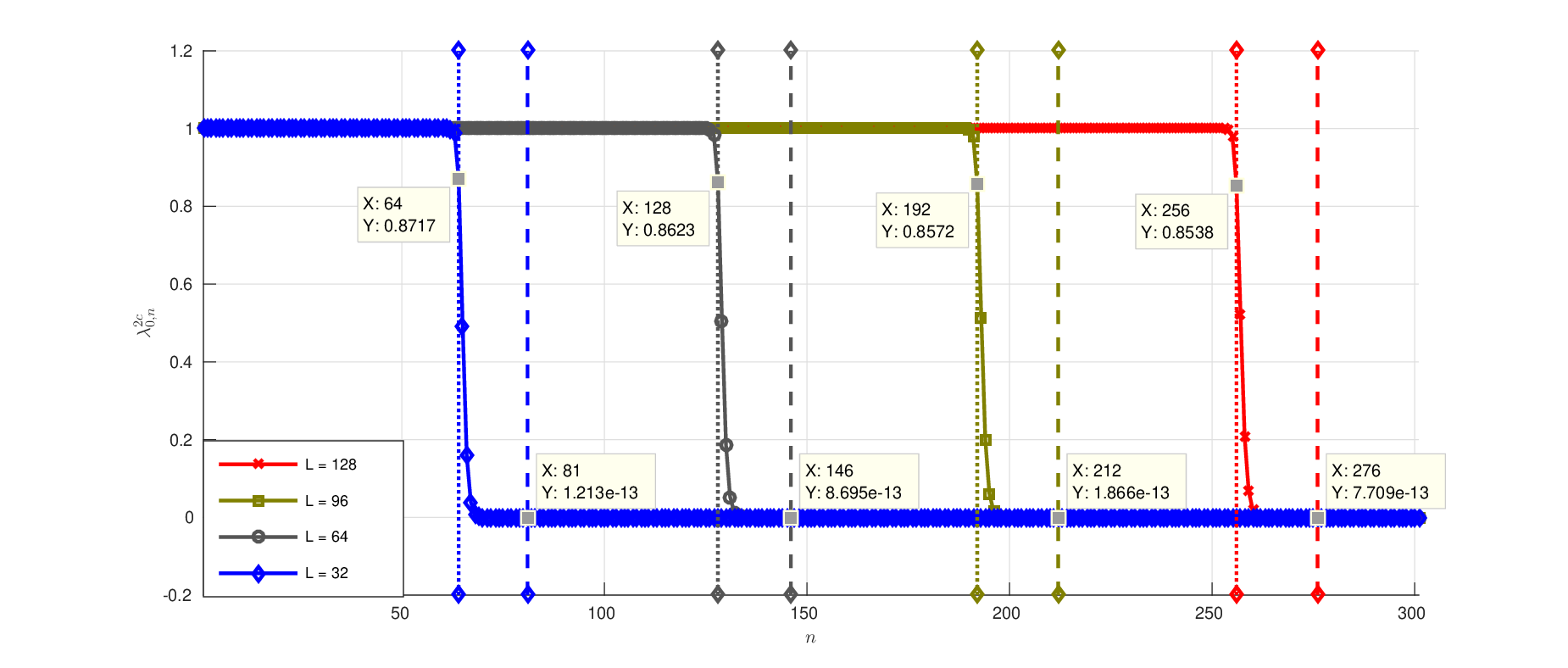}
	\caption[lambda_0_n spectrum]
    {Behaviour of $\left|\lambda_{0,n}^{2c}\right|$ for $c=\pi L$ (Nyquist sampling) and several values of $L$. For each value of $L$, the dotted vertical line represents $n={2c}/{\pi}$, and the dashed vertical line represents the first index $n$ for which $\left|\lambda_{0,n}^{2c}\right| \leq 10^{-12}$, which we denote by $n_1^{2c}$. First, we observe that ${2c}/{\pi}$ is the exact index for which $\left|\lambda_{0,n}^{2c}\right|$ begins its rapid decay, and second, it is evident that the difference between $n_1^{2c}$ and the estimate ${2c}/{\pi}$ grows extremely slowly with $L$, making the asymptotic estimate $n_1^{2c} \sim {2c}/{\pi}$ sufficiently accurate even for moderate values of $L$.} \label{fig:lambda_0_n spectrum}
\end{figure}

With regard to condition \eqref{eq:numerical integration radial nodes condition}, although it may seem somewhat daunting, the sum on the left hand-side is dominated by the values of $\left| \lambda_{0,k}^{2c} \right|$ and their decay properties. We argue in Appendix~\ref{appendix:Behevaior of lambda} that the number of non-negligible (relative to machine precision) values of $\left| \lambda_{0,k}^{2c} \right|$ is only about ${2c}/{\pi}$, that is
\begin{equation}
\left\vert \left\lbrace k: \; \left| \lambda_{0,k}^{2c} \right| > \epsilon \right\rbrace \right\vert \sim \frac{2c}{\pi}, \label{eq:lambda behaviour asymptotic}
\end{equation}
for some small $\epsilon$ and a sufficiently large $c$. This observation stems from the fact that the values of $\left| \lambda_{0,k}^{2c} \right|$ become arbitrarily small (and decay at a super-exponential rate) once $k$ reaches $2c/\pi+O(\log{c})$. We provide numerical evidence for this claim in Figure~\ref{fig:lambda_0_n spectrum}.

Therefore, it is evident that condition~\eqref{eq:numerical integration radial nodes condition} can be satisfied if $\left|\lambda_{0,k}^{2c}\right|$ is sufficiently small for $k>2\mathcal{N}_r$, which together with~\eqref{eq:lambda behaviour asymptotic} implies that for $c=\pi L$ (Nyquist sampling)
\begin{equation}
\mathcal{N}_r \sim \frac{c}{\pi} = L. \label{eq:number of radial nodes}
\end{equation}
In total, by~\eqref{eq:number of quad nodes 1} and~\eqref{eq:number of radial nodes}, the number of quadrature nodes for $c=\pi L$ is
\begin{equation}
\sum_{\ell=1}^{\mathcal{N}_r} \mathcal{\mathcal{N}_\theta^{\ell}} \sim \frac{e}{2\pi}c^2 = \frac{\pi e}{2} L^2 \approx 4.27 L^2, \label{eq:total number of quadrature nodes}
\end{equation}
which is only slightly larger than the number of sampling points in the unit square.

Let us now turn our attention to the computational complexity of computing $\tilde{a}_{N,n}$ of~\eqref{eq:expansion coeffs quadrature scheme}. The first step is to compute $\phi^c(\omega_{\ell,j}^{2c})$ from~\eqref{eq:Phi def} for all quadrature nodes $\left\lbrace \omega_{\ell,j}^{2c} \right\rbrace$, which can be implemented efficiently using the non-equispaced Fast Fourier Transform (NFFT)~\cite{dutt1993fast,greengard2004accelerating,potts2001fast,fessler2003nonuniform}. The computational complexity of such algorithms is
\begin{equation}
O\left( L^2\log{L} + \log{\epsilon^{-2}}P\right),
\end{equation}
where $L$ is the sampling rate, $\epsilon$ is the required accuracy of the transform, and $P$ is the number of evaluation points, which is equal to the number of quadrature nodes~\eqref{eq:total number of quadrature nodes}.
Next, we notice that for every value of $\ell$, the inner sum in~\eqref{eq:expansion coeffs quadrature scheme}
\begin{equation}
C_\ell^N \triangleq \sum_{j = 1}^{\mathcal{N}_\theta^{\ell}} \phi^c(\omega_{\ell,j}^{2c}) e^{- \frac{2\pi \imath j N}{\mathcal{N}_\theta^{\ell}}} \label{eq:coefficient eval inner sum}
\end{equation}
can be computed for multiple values of $N$ using the FFT in $O(\mathcal{N}_\theta^{\ell}\log{\mathcal{N}_\theta^{\ell}})$ operations, resulting in total of $O\left( \sum_{l=1}^{\mathcal{N}_r}\mathcal{N}_\theta^{\ell}\log{\mathcal{N}_\theta^{\ell}} \right)$ operations for all required values of $\ell$. By~\eqref{eq:total number of quadrature nodes}, this results in a computational complexity of $O(L^2\log{L})$ for $c=\pi L$. Lastly, the approximated expansion coefficients~\eqref{eq:expansion coeffs quadrature scheme} are given by the remaining outer sum
\begin{equation}
\tilde{a}_{N,n} = \sqrt{2\pi} \alpha_{N,n}^c \left(\frac{c}{2\pi L}\right)^2 \sum_{\ell = 1}^{\mathcal{N}_r} \tilde{\mathcal{W}}_{\ell}^{2c} \frac{R_{N,n}^c(r_\omega^\ell) r_\omega^\ell}{\mathcal{N}_\theta^{\ell}} C_\ell^N, \label{eq:coefficient eval outer sum}
\end{equation}
which can be computed for all indices $\left\lbrace N,n \right\rbrace \in \Omega_T$ in  $\mathcal{N}_r \left| \Omega_T \right|$ operations, where $\left| \Omega_T \right|$ denotes the cardinality of the index set $\Omega_T$.
From~\eqref{eq:Asymptotic number of terms} and~\eqref{eq:number of radial nodes}, the computational complexity of the last step is essentially $O(L^3)$, which thus governs the computational complexity of the entire procedure described in this section.
We end this discussion with the observation that the complexity of evaluating~\eqref{eq:coefficient eval outer sum} can be further reduced by exploiting the rapid decay of the functions $R_{N,n}^c(r)$ with the radius $r$ (see Figure~\ref{fig:pswfs illustration radial profile} for an illustration), due to which a substantial number of the terms involving $R_{N,n}^c(r_\omega^\ell)$ can be discarded for certain sets of the radii $\left\lbrace r_\omega^\ell \right\rbrace$ and indices $\left\lbrace (N,n) \right\rbrace$. To exemplify this point, for $T=1$, $L=150$ and $c=\pi L$, we have that about $30\%$ of the values $\left\lbrace \left|R_{N,n}^c(r_\omega^\ell)\right| \right\rbrace$ are below $10^{-12}$, and therefore can be safely discarded from~\eqref{eq:coefficient eval outer sum}.

\section{Steerable-PCA procedure} \label{section:Steerable PCA procedure}
As discussed in the introduction, steerable-PCA extends the classical PCA by artificially including in the analysed dataset all (infinitely many) planar rotations of each image.
In the previous sections, we have established a framework for expanding images using PSWFs, under the assumption
that the images to be approximated are sufficiently concentrated in space and frequency. Since PSWFs constitute an excellent basis for expanding images which are localized in space and frequency (see~\eqref{eq:Total approx err},~\eqref{eq:Total approx err 2} and~\cite{landa2016approximation}), we use them in this section to construct optimized basis functions for a given set of images and their rotations.

Let us suppose that our dataset consists of $M$ (sampled) images, where the $m$'th image is given by the samples $I_m\left(\frac{k}{L}\right)$ of some function $I_m(x) \in \mathcal{L}^2(\mathbf{D})$. We denote by $I_m^\varphi(x)$ the planar rotation of $I_{m}(x)$ by an angle $\varphi \in [0,2\pi)$.
Now, ideally, we would like to obtain basis functions that best approximate (in the sense of $\mathcal{L}^2(\mathbf{D})$) the images $I_m^\varphi(x)$ for all rotation angles $\varphi\in[0,2\pi)$. However, we do not have access to the underlying images $I_m(x)$, nor their rotations $I_m^\varphi(x)$. Therefore, we first replace them with their PSWFs-based approximations, denoted by $\hat{I}_m(x)$ (see~\eqref{eq:Coeff approx series}) and $\hat{I}^\varphi_m(x)$. Then, we derive a steerable-PCA procedure for the set of approximated images, and finally, we make the connection between the resulting steerable principal components of the approximated images and the original images.

The $k$'th principal component $g_k(x) \in \mathcal{L}^2(\mathbf{D})$ for the set of approximated images $\left \{\hat{I}_m^\varphi \right \}$, $m=0,\ldots,M-1$, is defined as
\begin{equation}
\begin{aligned}\label{eq:steerable PCA optimization}
g_k(x) = &\operatornamewithlimits{argmax}_{g(x)} \frac{1}{M}\sum_{m=0}^{M-1} \frac{1}{2\pi} \int_0^{2\pi} \left| \left\langle \hat{I}_m^\varphi (x) - \hat{\mu}(x), g(x) \right\rangle_{\mathcal{L}^2(\mathbf{D})} \right|^2 d\varphi, \\
 \text{s.t.} \;\; &\left\Vert g_k(x) \right\Vert_{\mathcal{L}^2(\mathbf{D})} = 1, \;\; \left\langle g_k(x),g_i(x) \right\rangle_{\mathcal{L}^2(\mathbf{D})} = 0, \;\; 0\leq i < k,
\end{aligned}
\end{equation}
where $\left\langle \cdot,\cdot \right\rangle_{\mathcal{L}^2(\mathbf{D})}$ denotes the standard inner product on $\mathcal{L}^2(\mathbf{D})$, and $\hat{\mu}(x)$ is the average of all approximated images and their planar rotations, given by
\begin{equation}
\hat{\mu}(x) = \frac{1}{M}\sum_{m=0}^{M-1} \frac{1}{2\pi} \int_0^{2\pi} \hat{I}_m^\varphi(x) d\varphi.
\end{equation}
In other words, the function $g_k(x)$ is expected to maximize the average projection norm (defined over all images and their rotations), such that $\left\lbrace g_k(x)\right\rbrace$ forms a set of orthonormal functions over $\mathcal{L}^2(\mathbf{D})$. The functions $\left\lbrace g_k(x) \right\rbrace$ are named the steerable principal components of our data set.
The formulation~\eqref{eq:steerable PCA optimization} differs from the classical formulation of PCA in the additional integration over the rotation angle $\varphi$, which has the interpretation of including all (infinitely many) rotations of all images in our data set.

After substituting~\eqref{eq:Coeff approx series} in~\eqref{eq:steerable PCA optimization} and exploiting the orthonormality and steerable structure of the PSWFs, the optimization problem~\eqref{eq:steerable PCA optimization} is reduced to a problem in the domain of the PSWFs expansion coefficients $\hat{a}_{N,n}$, for which the derivation in~\cite{zhao2013fourier} reveals that the solution is given by the eigenvectors of the $\left|\Omega_T\right|\times\left|\Omega_T\right|$  hermitian positive semi-definite matrix, whose entries are
\begin{equation}
C_{(N,n),(N^{'},n^{'})}=
\begin{cases}
\frac{1}{M}\sum_{m=0}^{M-1} \hat{c}_{N,n}^{m} \left({\hat{c}}_{N^{'},n^{'}}^{m}\right)^{*}, & N=N^{'},\\
0,  & N\neq N^{'},   \label{eq:rot cov mat explicit form}
\end{cases}
\end{equation}
where
\begin{equation}
\hat{c}_{N,n}^{m} =
\begin{cases}
\hat{a}_{N,n}^m - \frac{1}{M} \sum_{m'=0}^{M-1} \hat{a}_{0,n}^{m'} & N=0,\\
\hat{a}_{N,n}^m  & N \neq 0, \label{eq:b coeffs explicit form}
\end{cases}
\end{equation}
and $\hat{a}_{N,n}^{m}$, given by~\eqref{eq:Modified Prolates coefficient evaluation} (or~\eqref{eq:expansion coeffs quadrature scheme}), are the approximated expansion coefficients for the $m$'th image in our dataset, i.e. the coefficients of $\hat{I}_{m}$.
We refer to $C$ of~\eqref{eq:rot cov mat explicit form} as our rotationally-invariant covariance matrix (replacing the standard covariance matrix used in classical PCA).
Since the matrix $C$ enjoys a block-diagonal structure, we obtain its eigen-decomposition through the eigen-decomposition of its blocks. Let $\hat{\lambda}_1\geq\hat{\lambda}_2\geq\ldots\geq\hat{\lambda}_{\left|\Omega_T\right|}\geq 0$ be the eigenvalues of the matrix $C$ and let $\hat{g}_k$ be the eigenvector corresponding to eigenvalue $\hat{\lambda}_k$. For each pair $(\hat{\lambda}_{k},\hat{g}_{k})$, the entries of $\hat{g}_{k}$, denoted by $\hat{g}_{N,n}^k$, are nonzero only on some block of the matrix $C$ corresponding to an angular index $N_k$, and it can be shown that the functions $g_k(x)$ of~\eqref{eq:steerable PCA optimization} are recovered as
\begin{equation}
g_{k}(x) = \sum_{n=0}^{n_{k}} \hat{g}_{N_k,n}^k \psi_{N_{k},n} (x), \label{eq:sPC explicit form}
\end{equation}
where $n_{k}$ stands for the largest index $n$ such that $(N_{k},n) \in \Omega_T$. Equation~\eqref{eq:sPC explicit form}, together with~\eqref{eq:PSWFs complex form}, confirms that each $g_k(x)$ is a steerable function (as in~\cite{freeman1991design}).

By the formulation of~\eqref{eq:steerable PCA optimization}, it follows that the functions $g_k(x)$ form the optimal basis (in $\mathcal{L}^2(\mathbf{D})$) for expanding the approximated images $\left\lbrace \hat{I}_m(x) \right\rbrace$ and their rotations, such that if we define
\begin{equation}
\tilde{I}_{K,m}^{\varphi}(x) \triangleq \hat{\mu}(x) + \sum_{k=1}^{K} c_{k,m}^\varphi g_k(x), \qquad c_{k,m}^\varphi \triangleq e^{-\imath N_k \varphi} \sum_{n=0}^{n_k} \hat{c}_{{N_k},n}^m \left(\hat{g}_{N_k,n}^k\right)^* ,
\label{eq:steerable PCA expansion coeff}
\end{equation}
then we have that
\begin{equation}
\frac{1}{M}\sum_{m=0}^{M-1} \frac{1}{2\pi} \int_0^{2\pi} \left\Vert \hat{I}_m^\varphi(x) - \tilde{I}_{K,m}^{\varphi}(x) \right\Vert^2_{\mathcal{L}^2(\mathbf{D})} d\varphi \leq \sum_{k=K+1}^{\left|\Omega\right|} \hat{\lambda}_k, \label{eq:steerbale PCA approx error bound}
\end{equation}
where $\hat{\lambda}_k$ is the $k$'th eigenvalue of $C$.

Since the steerable principal components were computed for the set of approximated images $\left\lbrace \hat{I}_m(x) \right\rbrace$ and not for the set of underlying images $\left\lbrace {I}_m(x) \right\rbrace$, the bound in~\eqref{eq:steerbale PCA approx error bound} holds only for the set of the approximated images. Nonetheless, the PSWFs approximation scheme from Section~\ref{section:Image approximation based on PSWFs expansion} provides us with strict error bounds when expanding images localized in space and frequency. Therefore, when using~\eqref{eq:steerable PCA expansion coeff} to approximate our images $I_m(x)$, the error norm (averaged over all images and rotations) can be bounded by joining~\eqref{eq:steerbale PCA approx error bound} and~\eqref{eq:Total approx err 2}. In particular, for a truncation parameter $T$, we have that
\begin{equation}
\frac{1}{M}\sum_{m=0}^{M-1} \frac{1}{2\pi} \int_0^{2\pi} \left\Vert {I}_m^\varphi(x) - \tilde{I}_{K,m}^{\varphi}(x) \right\Vert^2_{\mathcal{L}^2(\mathbf{D})} d\varphi \leq \sum_{k=K+1}^{\left|\Omega_T\right|} \hat{\lambda}_k + 2\mathcal{E}(\varepsilon,\delta_c,T)\sqrt{\sum_{k=K+1}^{\left|\Omega_T\right|} \hat{\lambda}_k} + \mathcal{E}^2(\varepsilon,\delta_c,T), \label{eq:steerbale PCA approx error bound total}
\end{equation}
where $\tilde{I}_{K,m}^{\varphi}(x)$ is the expansion via the steerable principal components~\eqref{eq:steerable PCA expansion coeff}, and $\mathcal{E}(\varepsilon,\delta_c,T)$ is the approximation error term of~\eqref{eq:Total approx err 2} for the truncated series of PSWFs. In essence,~\eqref{eq:steerbale PCA approx error bound total} asserts that if the images are sufficiently localized in space and frequency, then for an appropriate truncation parameter $T$, the error in expanding $I_{m}(x)$ using the steerable principal components computed from $\hat{I}_{m}(x)$ is close to the smallest possible error given by~\eqref{eq:steerbale PCA approx error bound}.

\section{Algorithm summary and computational cost} \label{section:algorithm summary and computational cost}
We summarize the algorithms for evaluating the expansion coefficients $\hat{a}_{N,n}$ and $\tilde{a}_{N,n}$ (corresponding to the direct method from Section~\ref{section:Image approximation based on PSWFs expansion} and the efficient method of Section~\ref{section:Fast PSWFs coefficients approximation}, respectively) in Algorithms~\ref{alg:PSWFs epansion coeff eval simple} and \ref{alg:PSWFs epansion coeff eval efficient} respectively. The steerable-PCA procedure described in Section~\ref{section:Steerable PCA procedure} is summarized in Algorithm~\ref{alg:Steerable PCA procedure}.

\begin{algorithm}
\caption{Evaluating PSWFs expansion coefficients (direct method)}\label{alg:PSWFs epansion coeff eval simple}
\begin{algorithmic}[1]
\State{\textbf{Required:} An image $\left\lbrace I(\frac{k}{L}) \right\rbrace$ sampled on a Cartesian grid of size $(2L+1)\times(2L+1)$ with
$\frac{k}{L} \in \mathbf{Q}$, and $\mathbf{Q}=[-1,1]\times[-1,1]$.}
\State{\textbf{Precomputation:}
\begin{enumerate}
\item Choose a bandlimit $c$ ($\leq\pi L$) and a truncation parameter $T$.
\item Evaluate the PSWFs $\psi_{N,n}^c(\frac{k}{L})$ and their eigenvalues $\alpha_{N,n}^c$ according to~\cite{shkolnisky2007prolate} for $(N,n) \in \Omega_T$ (see~\eqref{eq:PSWFs truncation rule}), where $\frac{k}{L} \in \mathbf{D}$, and compute the normalized eigenvalues $\lambda_{N,n}^c = \frac{c}{2\pi} \alpha_{N,n}^c$.
\end{enumerate}
}
\State For all $(N,n)\in \Omega_T$, compute the expansion coefficients $\hat{a}_{N,n} =\nobreak \frac{\left\vert\lambda_{N,n}^{c}\right\vert^2}{L^2}  \underset{\frac{k}{L}\in\mathbf{D}}{\sum}I(\frac{k}{L})\left( {\psi}^c_{N,n}(\frac{k}{L})\right)^*$.
\end{algorithmic}
\end{algorithm}

\begin{algorithm}
\caption{Evaluating PSWFs expansion coefficients (efficient method)}\label{alg:PSWFs epansion coeff eval efficient}
\begin{algorithmic}[1]
\State{\textbf{Required:} An image $\left\lbrace I(\frac{k}{L}) \right\rbrace$ sampled on a Cartesian grid of size $(2L+1)\times(2L+1)$ with
$\frac{k}{L} \in \mathbf{Q}$, and $\mathbf{Q}=[-1,1]\times[-1,1]$.}
\State{\textbf{Precomputation:}
\begin{enumerate}
\item Choose a bandlimit $c$ ($\leq\pi L$) and a truncation parameter $T$.
\item Choose the number of radial nodes $\mathcal{N}_r$ and angular nodes $\mathcal{N}_\theta^{\ell}$ for $1 \leq \ell \leq \mathcal{N}_r$ (according to~\eqref{eq:numerical integration angular nodes condition} and~\eqref{eq:numerical integration radial nodes condition} or similar relaxed conditions).
\item Compute the quadrature nodes $\omega_{\ell,j}^{2c}=\left (r_{\omega}^{\ell}, \theta_{\omega}^{\ell,j}\right )$ and weights $\mathcal{W}_{\ell,j}^{2c}$ for $\ell=\nobreak 1,\ldots,\mathcal{N}_r$ and $j=1,\ldots,\mathcal{N}_\theta^{\ell}$, as described in~\cite{shkolnisky2007prolate}. \label{step:NFFT}
\item Evaluate the radial part of the PSWFs $R_{N,n}^c(r)$ at the radial quadrature nodes $r_\omega^\ell$
for $(N,n) \in \Omega_T$ (see~\eqref{eq:PSWFs truncation rule}). \label{step:numerical integration}
\end{enumerate}
}
\State Compute $\phi^c(\omega_{\ell,j}^{2c})$ from~\eqref{eq:Phi def} by NFFT.
\State For $(N,n)\in \Omega_T$, compute the expansion coefficients $\tilde{a}_{N,n}$ via equations~\eqref{eq:coefficient eval inner sum} and~\eqref{eq:coefficient eval outer sum}.
\end{algorithmic}
\end{algorithm}

\begin{algorithm}
\caption{PSWFs-based steerable-PCA}\label{alg:Steerable PCA procedure}
\begin{algorithmic}[1]
\State{\textbf{Required:} PSWFs expansion coefficients of $M$ images $\left\lbrace \hat{a}_{N,n}^m \right\rbrace_{m=0}^{M-1}$ for $(N,n)\in\Omega_T$ and a bandlimit~$c$.}
\State{\textbf{Precomputation:}} Evaluate the PSWFs $\psi_{N,n}^c(\frac{k}{L})$, $\frac{k}{L} \in \mathbf{D}$,  and their eigenvalues $\alpha_{N,n}^c$ for $(N,n) \in \Omega_T$ according to~\cite{shkolnisky2007prolate}.
\State Compute the expansion coefficients of the mean image $\hat{\mu}_{0,n} = \frac{1}{M}\sum_{m=0}^{M-1}\hat{a}_{0,n}^m$ for $n=0,\ldots,n_0$ where $n_0$ is the largest index $n$ such that $(0,n) \in \Omega_T$.
\State Update $\hat{a}_{0,n}^m \gets \hat{a}_{0,n}^m - \hat{\mu}_{0,n}$.
\State Compute the eigenvalues $\hat{\lambda}_1,\ldots,\hat{\lambda}_{\left|\Omega_T\right|}$ and eigenvectors $\hat{g}_1,\ldots,\hat{g}_{\left|\Omega_T\right|}$ of the matrix $C$ (from~\eqref{eq:rot cov mat explicit form}) by diagonalizing each of its blocks separately. \label{step:ev}
\State Compute the sampled basis functions $g_\ell(\frac{k}{L})=\sum_{n=0}^{n_{\ell}} \hat{g}_{N_\ell,n}^\ell {\psi}_{N_\ell,n}^c(\frac{k}{L})$,
where $n_{\ell}$ stands for the largest index $n$ such that $(N_{\ell},n) \in \Omega_T$, and $\hat{g}_{N_\ell,n}^\ell$ are the entries of the eigenvector $\hat{g}_{\ell}$ corresponding to the pair $(N_\ell,n)$. \label{step:compute gk}
\State Compute the coefficients of $I_{m}$ in the steerable-PCA basis by $c_{\ell,m}= \sum_{n=0}^{n_\ell} \hat{a}_{{N_\ell},n}^m \left(\hat{g}_{N_\ell,n}^\ell\right)^*$.\label{step:compute dk}
\end{algorithmic}
\end{algorithm}

We now turn our attention to the computational complexity of Algorithms~\ref{alg:PSWFs epansion coeff eval simple}, \ref{alg:PSWFs epansion coeff eval efficient}, and~\ref{alg:Steerable PCA procedure}. We omit the pre-computation steps from the complexity analysis as they can be performed only once per setup, and do not depend on the specific images.

Since we have $O(L^2)$ PSWFs in our expansions (corresponding to the indices in the set $\Omega_T$) and $O(L^2)$ equally-spaced Cartesian samples in the unit disk, computing the expansion coefficients of $M$ images using the direct approach of Algorithm~\ref{alg:PSWFs epansion coeff eval simple} requires $O(ML^4)$ operations. On the other hand, Algorithm~\ref{alg:PSWFs epansion coeff eval efficient} allows us to obtain the expansion coefficients in $O(ML^3)$ operations, since the NFFT in step~\ref{step:NFFT} requires $O(L^2\log{L})$ operations and step~\ref{step:numerical integration} can be implemented using $O(L^2\log{L} + L^3)$ operations, for each image.

Although Algorithm~\ref{alg:PSWFs epansion coeff eval efficient} and the method described in~\cite{zhao2014fast} (based on Fourier-Bessel basis functions) have the same order of computational complexity, our method enjoys a twofold asymptotic speedup in the coefficients' evaluation, and often it runs about three/four times faster. The twofold asymptotic speedup is because we need asymptotically only half the number of radial nodes for the numerical integration (see analysis in Section \ref{section:Fast PSWFs coefficients approximation} and Appendix~\ref{appendix:Behevaior of lambda}) compared to the Gaussian quadratures used in~\cite{zhao2014fast}, which dictates the constant in the leading term of the computational complexity. The greater speedup observed in practice stems from the fact that practically, the most time-consuming operation in both methods is the NFFT, which depends heavily on the total number of quadrature nodes. In our integration scheme, the total number of quadrature nodes is about $1/4$ of the number of nodes in~\cite{zhao2014fast}.

Next, as of the computational complexity of the steerable-PCA (Algorithm~\ref{alg:Steerable PCA procedure}), forming the  blocks of the matrix $C$ requires $O(ML^3)$ operations, and then the eigen-decomposition of each block requires $O(L^3)$ operations, where there are $O(L)$ different blocks in the matrix. Therefore, obtaining the eigenvalues and eigenvectors in step~\ref{step:ev} of Algorithm~\ref{alg:Steerable PCA procedure} requires $O(ML^3+L^4)$ operations. As pointed out in~\cite{zhao2013fourier} and~\cite{zhao2014fast}, the eigenvalues and eigenvectors can be also obtained from the SVD of the matrices of coefficients $\hat{c}_{N,n}^m$ from which the blocks of $C$ are obtained.

Following~\eqref{eq:sPC explicit form}, if we have $O(L^2)$ basis functions $g_{k}(x)$, their evaluation on the Cartesian grid requires $O(L^5)$ operations. Sometimes it may be more convenient to evaluate these basis functions on a polar grid instead, in which case the computational complexity reduces to $O(L^4)$.

Lastly, computing the expansion coefficients of the images in the steerable basis via~\eqref{eq:steerable PCA expansion coeff} requires $O(ML^3)$ operations, since $O(L)$ operations are required to compute a single expansion coefficient for every image.

We point out though, that often only a small fraction of the basis functions are chosen for subsequent processing (via their eigenvalues), so the contribution of steps~\ref{step:compute gk} and~\ref{step:compute dk} in Algorithm~\ref{alg:Steerable PCA procedure} to the overall running time of the steerable-PCA procedure is usually negligible.

To summarize, the computational complexity of the entire procedure (computing PSWFs expansion coefficients + steerable-PCA), when using Algorithm~\ref{alg:PSWFs epansion coeff eval efficient} to compute the expansion coefficients, is $O(ML^3+L^5)$ when sampling the steerable principal components on the Cartesian grid, and $O(ML^3+L^4)$  when sampling them on a polar grid.
It is important to note that, although Algorithm~\ref{alg:PSWFs epansion coeff eval simple} for evaluating PSWFs expansion coefficients suffers from an inferior order of computational complexity (compared to Algorithm~\ref{alg:PSWFs epansion coeff eval efficient}), it is simpler to implement and may still run faster (due to optimized implementations of the scalar product on CPUs and GPUs), particularly for small values of $L$.

\section{Steerable-PCA in the presence of noise} \label{section:transformation spectrum}
Up to this point, we have presented a method for computing the steerable-PCA of a set of images localized in space and frequency, sampled on a Cartesian grid.
In many practical settings however, the images are corrupted with noise. Therefore, it is beneficial to understand the impact this noise has on the PSWFs expansion coefficients, and in particular, it is generally convenient if the transformation to the expansion coefficients does not alter the spectrum of the noise. In this section, we demonstrate numerically that for sufficiently localized images in space/frequency, for which we can choose a truncation parameter $T\gg 1$ (see~\eqref{eq:PSWFs truncation rule}), the transformation to the PSWFs expansion coefficients is essentially orthonormal. In particular, the higher the truncation parameter $T$, the closer is our steerable-PCA to orthonormality.

\begin{figure}
  \centering
  \subfloat[$T=10^{6}$]{
    \includegraphics[width=0.5\textwidth]{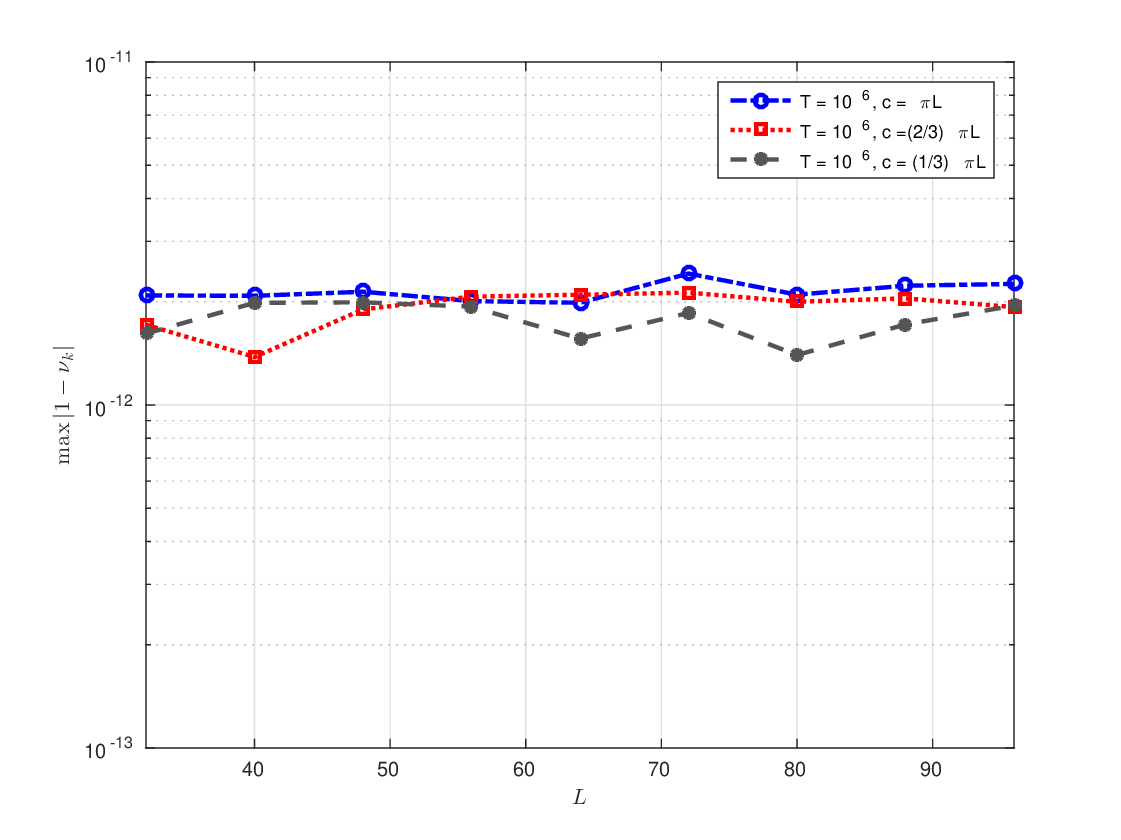}
    }
    \subfloat[$T=10^3$]{
    \includegraphics[width=0.5\textwidth]{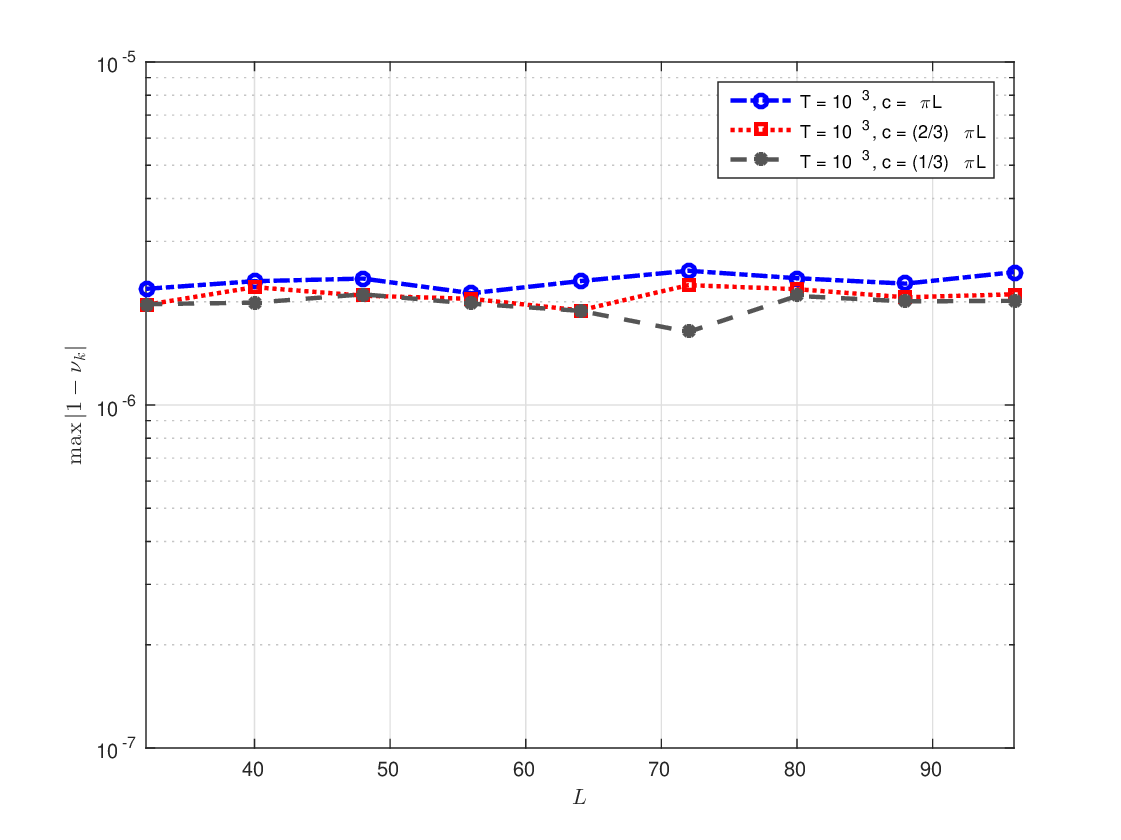}
    }
	\caption[Deviation from orthonormality]
    {Measured deviation of the eigenvalues of $H_c$ from 1 for different values of the truncation parameter $T$, the bandlimit $c$, and the sampling resolution $L$. We notice that the deviation from orthogonality remains approximately constant for different values of $L$ and $c$. Specifically, for $T=10^6$ we notice that $H_c$ is practically orthogonal.} \label{fig:PSWF spectrum}
\end{figure}

\begin{figure}
  \centering
    \includegraphics[width=0.6\textwidth]{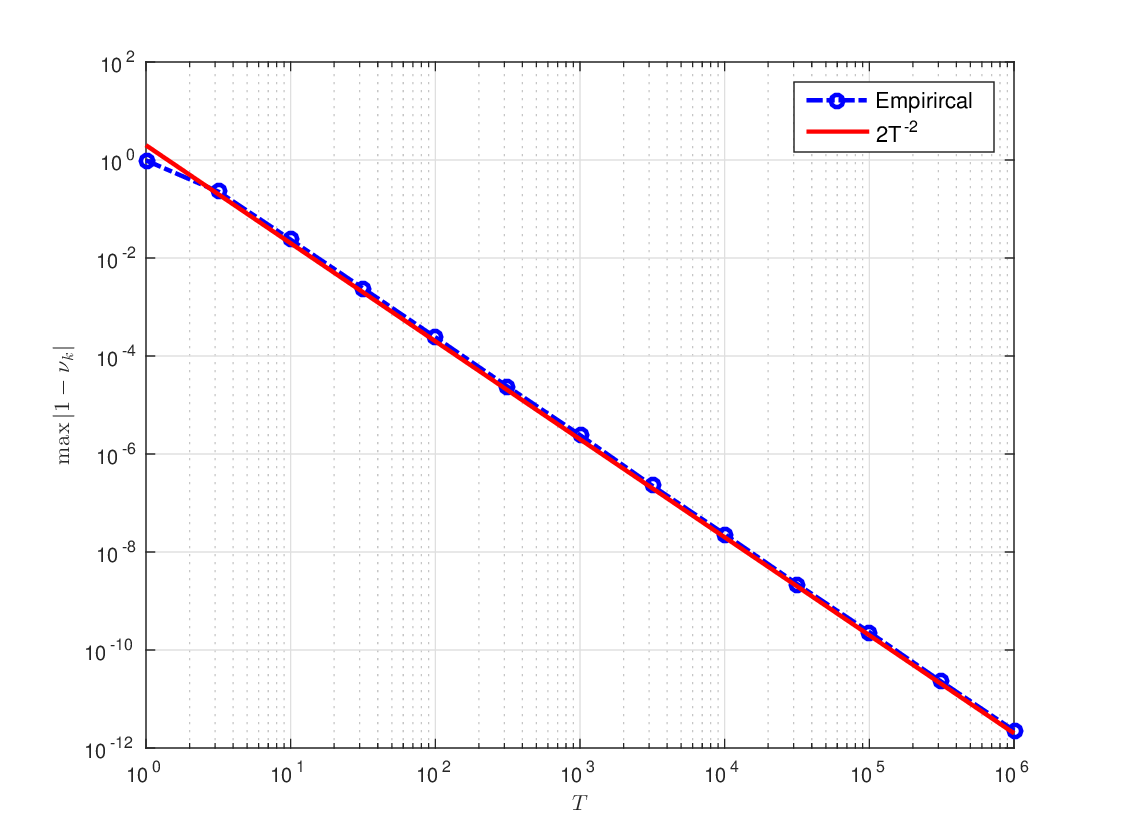}
	\caption[Deviation from orthonormality by T]
    {Measured deviation of the eigenvalues of $H_c$ from 1 as a function of $T$, versus the function $2T^{-2}$, for $L=96$ and $c=\pi L$. We obsereve that the truncation parameter $T$ gives us direct control over the spectrum of $H_c$ with $\max_{k}\left| 1 - \nu_k \right| \approx 2T^{-2}$, for $T \geq 10$.} \label{fig:PSWF spectrum as function of T}
\end{figure}

Let us denote by $I$ a column vector consisting of the clean (Cartesian) samples of an input image. Suppose that our image is corrupted by an additive noise such that
\begin{equation}
\tilde{I} = I + \xi, \label{eq:additive noise}
\end{equation}
where $\xi$ is zero mean noise vector with covariance matrix $R_\xi$.
From~\eqref{eq:Modified Prolates coefficient evaluation}, we can compute our approximated expansion coefficients by
\begin{equation}
\hat{a} = \left( \hat{\psi}^c \right)^* \tilde{I} = \left( \hat{\psi}^c \right)^* I + \left( \hat{\psi}^c \right)^* \xi,
\end{equation}
where the operator $(\cdot)^*$ stands for the conjugate-transpose, and $\hat{\psi}^c$ denotes the matrix whose columns contain samples of $\psi_{N,n}^c(x) \left\vert\lambda_{N,n}^{c}\right\vert^2 / L^2$ inside the unit disk, with different columns corresponding to different pairs of indices $(N,n)$.
If we define the vector of additive noise in the expansion coefficients as
\begin{equation}
\hat{\xi} = \left( \hat{\psi}^c \right)^* \xi,
\end{equation}
then its covariance matrix is provided by
\begin{equation}
R_{\hat{\xi}} = {\left( \hat{\psi}^c \right)}^* R_\xi \hat{\psi}^c.
\end{equation}
Now, if the noise in~\eqref{eq:additive noise} is white ($R_\xi = \sigma^2 I$), and in order to preserve the covariance of the noise, we would require the matrix
\begin{equation}\label{eq:Hc}
H_c \triangleq {\left( \hat{\psi}^c \right)}^* \hat{\psi}^c
\end{equation}
to be as close as possible to the identity matrix, or equivalently, that the eigenvalues of the matrix $H_c$ are as close as possible to $1$.
As the matrix $H_c$ is hermitian and positive semi-definite, it has non-negative real-valued eigenvalues $\nu_1,\nu_2,\ldots,\nu_{\Omega_T}$. To determine how close $H_{c}$ is to the identity matrix, we evaluated numerically the maximal distance (in absolute value) between the eigenvalues $\left\lbrace \nu_k \right\rbrace$ and $1$, and the results are plotted in Figure~\ref{fig:PSWF spectrum} for various values of $L$, $T$ and the bandlimit $c$.
It is noteworthy that for $T=10^6$ the eigenvalues of $H_c$ differ from $1$ by about $10^{-12}$. We have also observed that the spectrum of the matrix $H_c$ is related to the truncation parameter $T$ by
\begin{equation}
\max_{k}\left| 1 - \nu_k \right| \approx 2 T^{-2}, \label{eq:PSWFs samples orthogonality}
\end{equation}
for $T \geq 10$. This observation is exemplified numerically in Figure~\ref{fig:PSWF spectrum as function of T}.
As the matrix $H_c$ in~\eqref{eq:Hc} is essentially orthogonal for sufficiently large values of $T$, it is important to mention that if the images are sufficiently localized in space and frequency, it is often possible to choose the value of $T$ as to enjoy the orthogonality of the transform, while keeping the bound in~\eqref{eq:Total approx err 2} sufficiently small.

\section{Numerical experiments} \label{section:Numerical experiments}
We begin by demonstrating the running times of our algorithm for large datasets. We generated several sets of $20,000$ white noise images, where the sets consist of images with different $L$ for each set, and compared the running time of our algorithm with the FFBsPCA algorithm of~\cite{zhao2014fast}, where the coefficients' evaluation for our method was carried out using both Algorithm~\ref{alg:PSWFs epansion coeff eval simple} (direct method) and Algorithm~\ref{alg:PSWFs epansion coeff eval efficient} (efficient method). All of the algorithms were implemented in Matlab, and were executed on a dual Intel Xeon X5560 CPU (8 cores in total), with 96GB of RAM running Linux.
Whenever possible, all $8$ cores were used simultaneously, either explicitly using Matlab's \texttt{parfor}, or implicitly, by employing Matlab's implementation of BLAS, which takes advantage of multi-core computing. As for the NFFT implementation, we used the software package~\cite{keiner2009using}, with an oversampling of $2$ and a truncation parameter $m=6$ (which provides accuracy close to machine precision). The running times (in seconds) are shown in Table~\ref{tbl:run time comparison}.
\begin{table}
\centering
\begin{tabular}{cccc}
\hline \rule{0pt}{4ex} $L$   & \MyHead{3cm}{PSWFs-direct \\ (Algs.~\ref{alg:PSWFs epansion coeff eval simple}+\ref{alg:Steerable PCA procedure})} & \MyHead{3cm}{PSWFs-efficient \\ (Algs.~\ref{alg:PSWFs epansion coeff eval efficient}+\ref{alg:Steerable PCA procedure})} & FFBsPCA~\cite{zhao2014fast} \\ \hline
32 & 6 & 25 & 47 \\
64 & 95 & 59 & 151 \\
96 & 491 & 117 & 363 \\
128 & 1625 & 204 & 697 \\\hline
\end{tabular}
\caption{Comparison of algorithms' running times, for $20,000$ images consisting of white noise, for $T=10$ and several values of $L$. All timings are given in seconds.} \label{tbl:run time comparison}
\end{table}
As anticipated, Algorithm~\ref{alg:PSWFs epansion coeff eval simple} runs faster than Algorithm~\ref{alg:PSWFs epansion coeff eval efficient} for small image sizes, but becomes significantly slower for larger values of $L$.
As for the running time of our algorithm versus FFBsPCA~\cite{zhao2014fast}, we have mentioned in Section~\ref{section:algorithm summary and computational cost} that our algorithm is asymptotically two times faster, since the number of radial nodes in our quadrature scheme is asymptotically half that of FFBsPCA. In addition, the total number of quadrature nodes in our scheme is about a quarter of that of FFBsPCA, and since the NFFT procedure for evaluating the Fourier transform of the sampled images on a polar grid (see~\eqref{eq:Phi def}) is the most time consuming step of both algorithms, it is expected that our algorithm will be faster than FFBsPCA by a factor between $2$ and $4$, which indeed agrees with the results in Table~\ref{tbl:run time comparison}. In all scenarios tested, most of the computation time was spent on evaluating the PSWFs expansion coefficients (Algorithm~\ref{alg:PSWFs epansion coeff eval efficient}), and only a small fraction of the time on the eigen-decomposition of the rotationally-invariant covariance matrix. Table~\ref{tbl:inner time comparison} summarizes the time spent on the evaluation of the PSWFs expansion coefficients (using Algorithm~\ref{alg:PSWFs epansion coeff eval efficient}) versus time spent on the eigen-decomposition of the rotationally-invariant covariance matrix.

\begin{table}
\centering
\begin{tabular}{ccc}
\hline \rule{0pt}{4ex} $L$   & \MyHead{4cm}{PSWFs coefficients \\ evaluation (Alg.~\ref{alg:PSWFs epansion coeff eval efficient})} & \MyHead{4cm}{Eigen-decomposition \\ (Alg.~\ref{alg:Steerable PCA procedure} steps 3-6)} \\\hline
32 & 24.5 & 0.5 \\
64 & 56.5 & 2.5 \\
96 & 111 & 6 \\
128 & 193 & 11 \\ \hline
\end{tabular}
\caption{Running times of coefficients' evaluation and eigen-decomposition for the efficient PSWFs-based method. All timings are given in seconds.} \label{tbl:inner time comparison}
\end{table}
Next, we demonstrate our algorithms on simulated single-particle cryo-electron microscopy (cryo-EM) projection images. In single-particle cryo-EM, one is interested in reconstructing a three-dimensional model of a macromolecule (such as a protein) from its two-dimensional projection images taken by an electron microscope. The procedure begins by embedding many copies of the macromolecule in a thin layer of ice (hence the 'cryo' in the name of the procedure), where due to the experimental setup the different copies are frozen at random unknown orientations. Then, an electron microscope acquires two-dimensional projection images of the electron densities of these macromolecules. This procedure of image acquisition can be modelled mathematically as the Radon transform of a volume function evaluated at random viewing directions. Due to the properties of the imaging procedure, each projection image generated by the electron microscope undergoes a convolution with a kernel, referred to as the ``contrast transfer function'' (CTF), which is known to have a Gaussian  envelope~\cite{Frank}. Since the unknown volume is essentially limited in space, and since the behaviour of the CTF dictates that all images are localized in Fourier domain, we conclude that projection images obtained by single-particle cryo-EM are essentially limited to circular domains in both space and frequency.
We emphasize that although the goal in single particle cryo-EM is to reconstruct the three-dimensional structure of the macromolecule, the input to the reconstruction process is only the set of two-dimensional images. Now, since the in-plane rotation of each single-particle cryo-EM image in the detector plane is arbitrary and irrelevant for the reconstruction, the image processing methods applied to the input image dataset, such as denoising and classification, should be invariant to these in-plane rotations. This observation explains why these images are suitable for exemplifying our steerable-PCA algorithm.

\begin{figure}
  \centering
  	\subfloat[]{
    \includegraphics[width=0.3\textwidth]{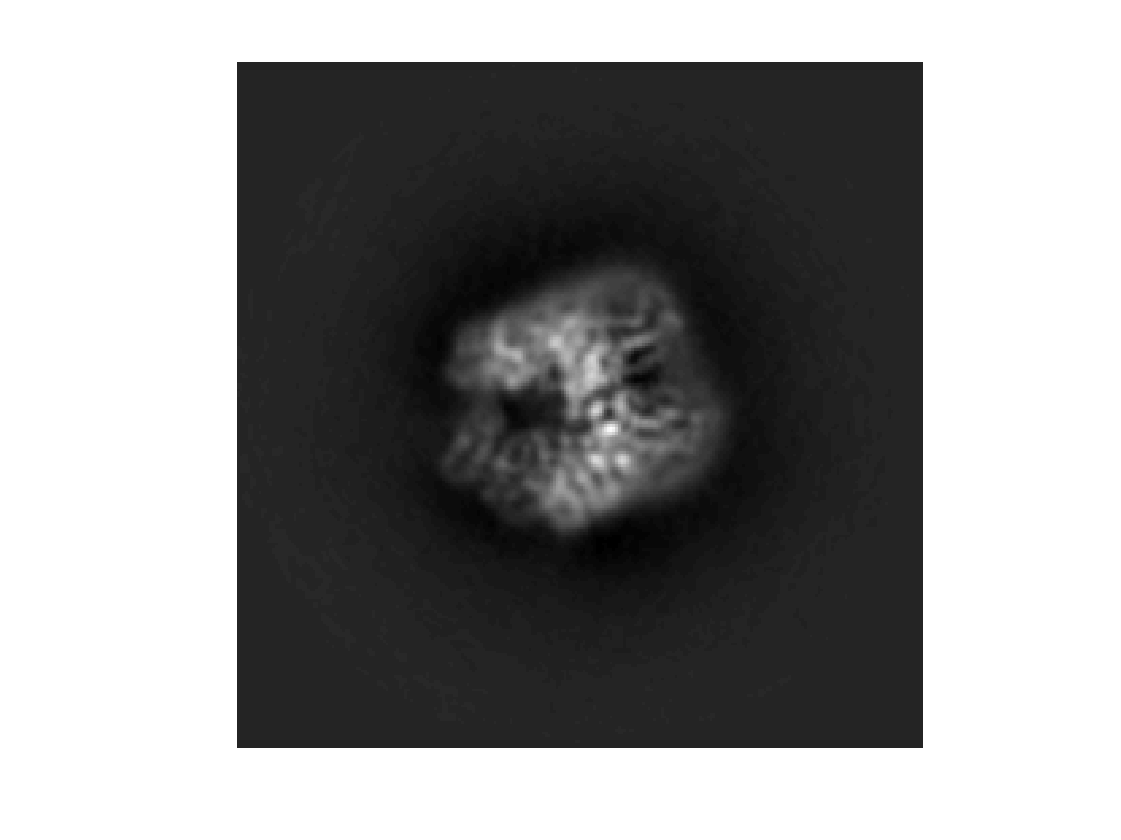}
    }
    \subfloat[]{
    \includegraphics[width=0.3\textwidth]{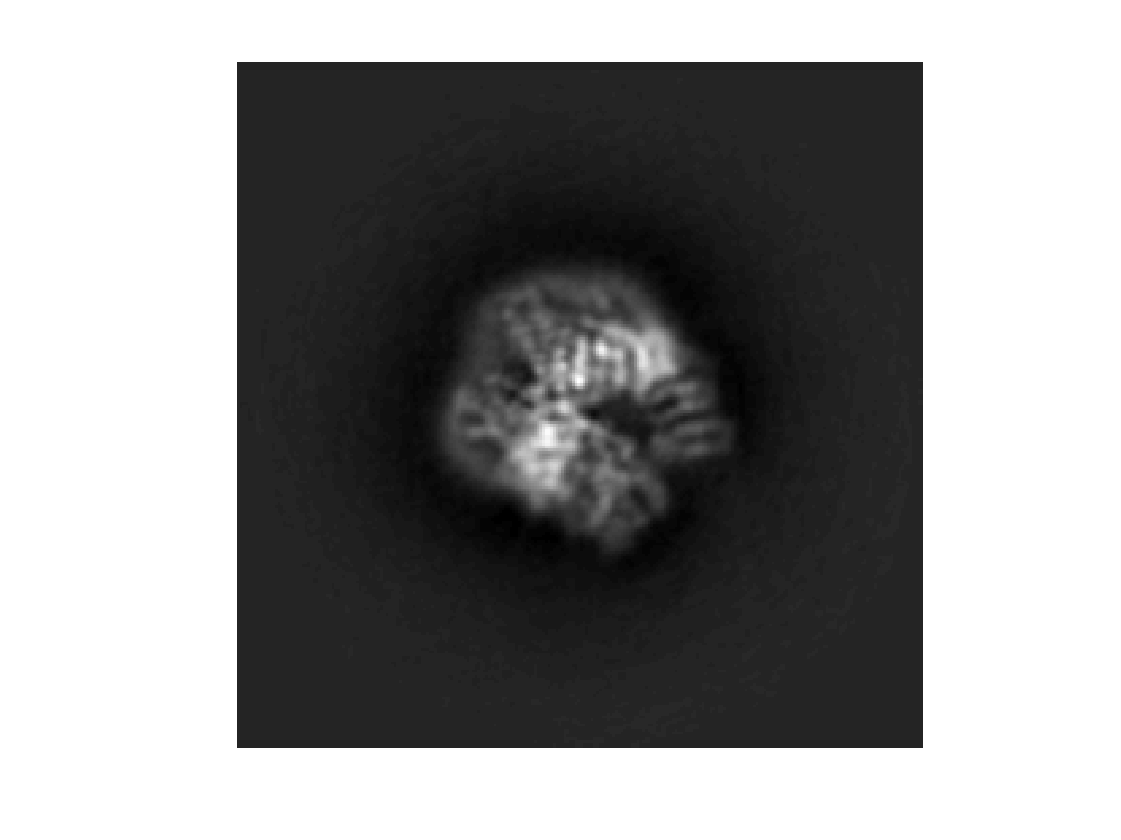}
    }
    \subfloat[]{
    \includegraphics[width=0.3\textwidth]{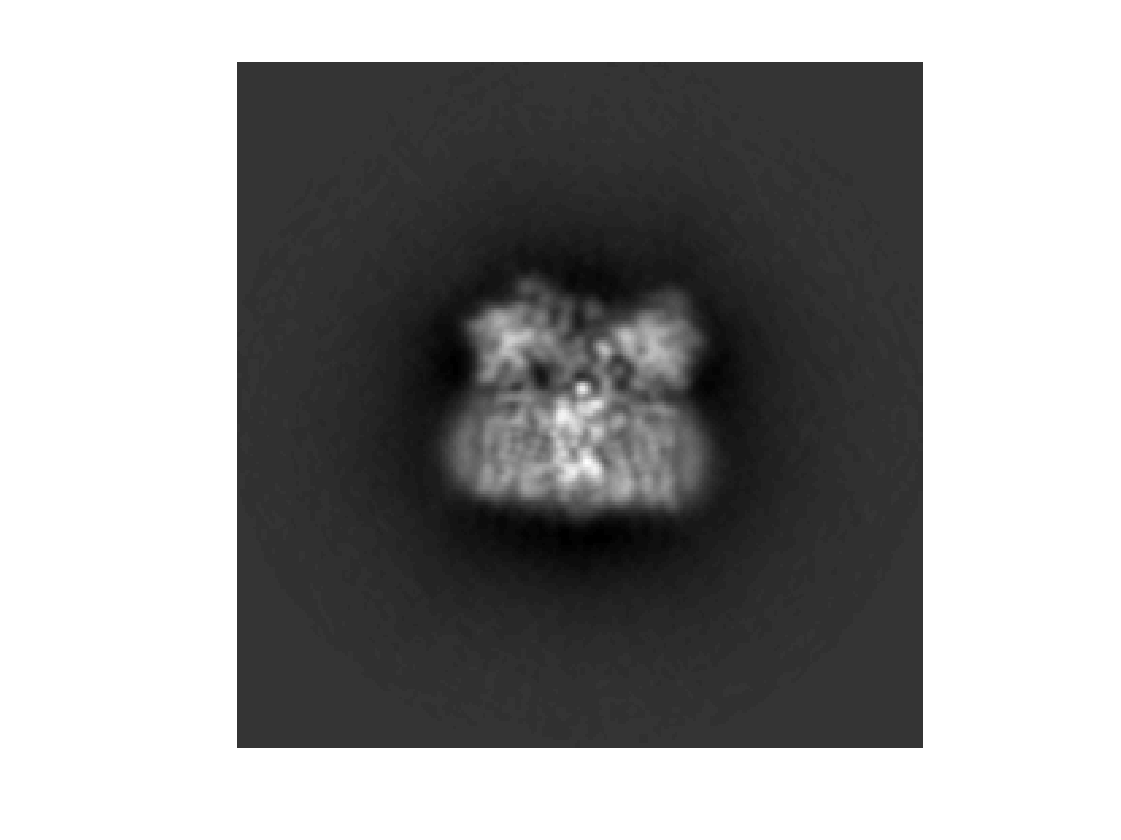}
    }
	\caption[typical_projection_images]
	{Sample of 3 simulated noiseless projection images at a resolution of $L=127$ pixels.} \label{fig:projection images}
\end{figure}

\begin{figure}
  \centering
    \includegraphics[width=0.65\textwidth]{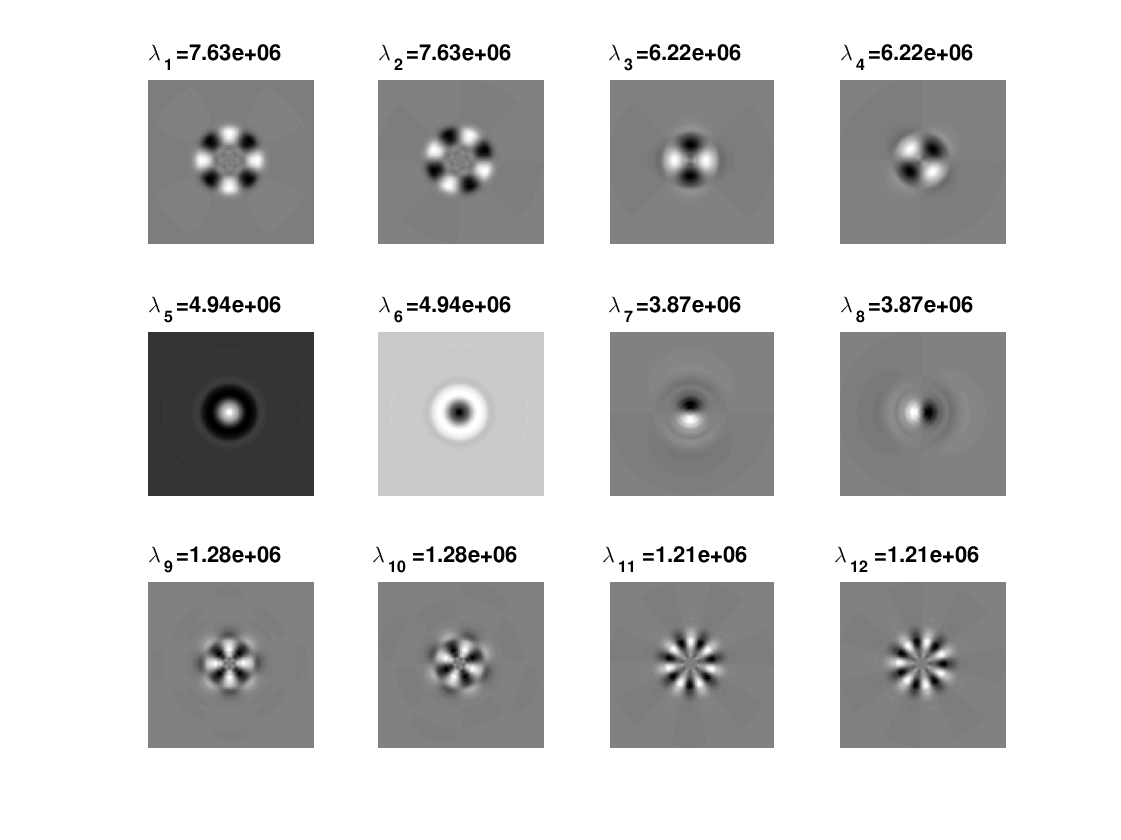}
	\caption[sPCA_eigenfunctions]
	{The first $12$ eigenfunctions with largest eigenvalues, obtained for $T=10^{-3}$. } \label{fig:sPCA eigenfunctions}
\end{figure}

\begin{figure}
  \centering    	
  	\subfloat[Comparison of theoretical and empirical errors]{	
    \includegraphics[width=0.65\textwidth]{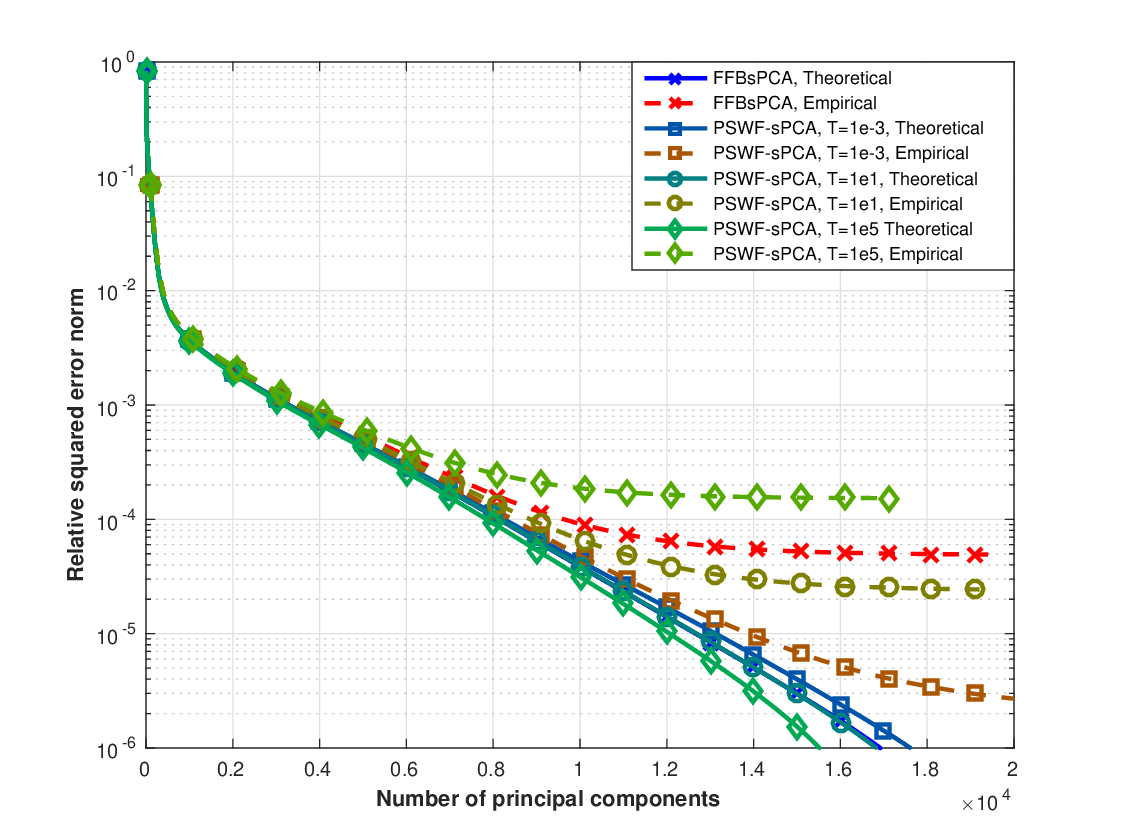} \label{fig:sPCA approx error}
    }  \\
    \subfloat[Approximation quality by different bases]{
    \includegraphics[width=0.65\textwidth]{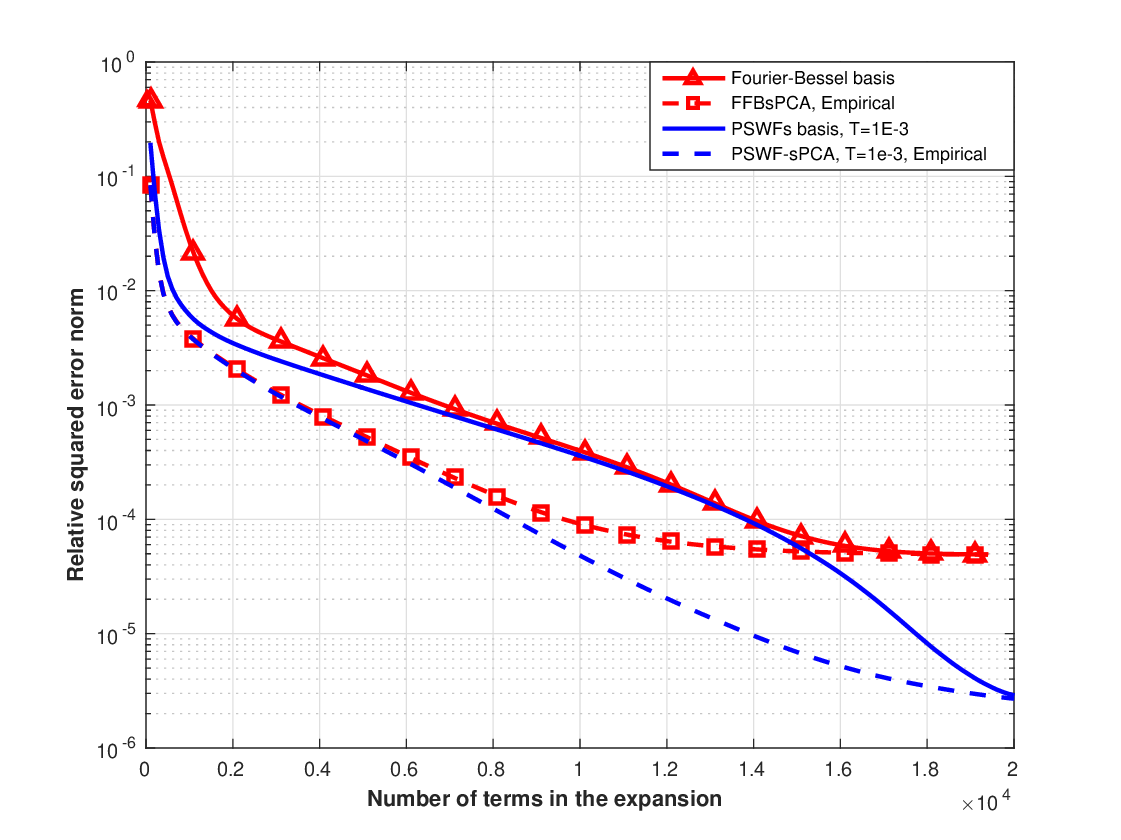} \label{fig:FB vs PSWF vs sPCA}
    }
	\caption[sPCA_approx_error]
	{The ratio between the squared error norm and the squared norm of the images, when expanding the images with different numbers of basis functions.
	Figure~\ref{fig:sPCA approx error} compares theoretical errors with empirical errors for our PSWFs-based algorithm (for several values of $T$) and the FFBsPCA algorithm of~\cite{zhao2014fast}. Curves which correspond to theoretical errors are obtained from the residual eigenvalues of the steerable-PCA procedure (see~\eqref{eq:steerbale PCA approx error bound}), and empirical errors correspond to measured errors between the reconstructed and original projection images. Figure~\ref{fig:FB vs PSWF vs sPCA} illustrates the errors due to steerable-PCA expansions, PSWFs expansions and Fourier-Bessel expansions. Note that the $x$-axis, which counts the number of basis functions used in each expansion, counts only basis functions with non-negative angular indices (since these are sufficient for encoding real-valued images).}
\end{figure}

To demonstrate the accuracy of our method, we simulated $10,000$ clean projection images from a noiseless three-dimensional density map (EMD-5578 from The Electron Microscopy Data Bank (EMDB)~\cite{Lawson04012016}) using the ASPIRE package~\cite{aspire}, and obtained steerable principal components using both our fast algorithm and FFBsPCA~\cite{zhao2014fast}. Then, we used different numbers of principal components to reconstruct the images, and compared the theoretical error predicted by the residual eigenvalues ({see}~\eqref{eq:steerbale PCA approx error bound}) with the empirical error obtained by comparing the reconstructions to the original images. Obviously, the difference between the two errors is due to the error incurred by the images' expansion scheme (PSWFs or Fourier-Bessel functions). Typical projection images from this dataset can be seen in Figure~\ref{fig:projection images}. To simplify the setting and the exposition, we used the projection images as they were obtained from the given volume, and we did not crop, filter, or process them beforehand. Therefore, we assumed a bandlimit $c=\pi L$ throughout the experiment. We note that in general it is possible to crop and filter the images (i.e. choose a smaller bandlimit~$c$) without significantly degrading the quality of the images (up to some prescribed accuracy), thus reducing the computational burden of the algorithms. This can be accomplished either by power density estimation (as demonstrated in~\cite{zhao2014fast}), or by employing more sophisticated and dataset-specific estimation techniques for the B-factor (which governs the Gaussian envelope decay of the CTF), see for example~\cite{mallick2005ace}. Figure~\ref{fig:sPCA eigenfunctions} depicts the first $12$ eigenfunctions obtained by our steerable-PCA algorithm for this dataset. In Figure~\ref{fig:sPCA approx error} we show the relative error norms for the FFBsPCA algorithm and our PSWFs-based algorithm, using several values of $T$ and with different numbers of eigenfunctions in the reconstruction. As expected, the theoretical and empirical relative error norms coincide when a small number of principal components is used in the reconstruction, yet when a large number of principal components is used, the error incurred by the expansion of the images (either using PSWFs or Fourier-Bessel) comes into play and dominates the overall error. As guaranteed by~\eqref{eq:steerbale PCA approx error bound total} and~\eqref{eq:Total approx err 2} for space/frequency localized images, smaller values of $T$ lead to smaller approximation errors, and we notice that our PSWFs-based method outperforms the FFBsPCA algorithm in terms of accuracy for $T=10$ and $T=10^{-3}$. It is also important to mention that the number of PSWFs taking part in the approximation of each image does not increase significantly when lowering $T$  (see~\eqref{eq:Asymptotic number of terms}). On the other hand, if one allows the approximation error to be of the order of $10^{-3}$ to $10^{-4}$, then it is also possible to use a very large truncation parameter, such as $T=10^5$, for which the PSWFs-based transform enjoys superior orthogonality properties (see Figure~\ref{fig:PSWF spectrum}), as well as shorter expansions. Often, such scenarios arise when handling noisy datasets. In a way, the truncation parameter provides us with flexibility to adapt the approximation scheme to the specific setting. In Figure~\ref{fig:FB vs PSWF vs sPCA} we show the approximation errors obtained by expanding the images in the PSWFs basis and in the Fourier-Bessel basis (without the rotationally-invariant orthogonalization), where we sorted the basis functions according to their contribution to the expansion. It is evident that PSWFs are more adapted to expanding our image dataset than Fourier-Bessel, which is reasonable due to the underlying model behind these images, whereas both bases provide lower accuracy than the steerable principal components obtained from our PSWFs-based steerable-PCA.

\section{Summary and discussion} \label{section:Summary and discussion}
In this paper, we utilized PSWFs-based computational tools to construct fast and accurate algorithms for obtaining steerable principal components of large image datasets. The accuracy of our algorithms are guaranteed under the assumptions of localization of the images in space and frequency, which are natural assumptions for many datasets, particularly in the field of tomography. For $M$ images, each sampled on a Cartesian grid of size $(2L+1)\times (2L+1)$, the computational complexity for obtaining the steerable principal components is $O(M L^3)$ operations, and their accuracy is of the order of the localization of the images in space and frequency, i.e. the norm of the images outside the unit disk in space, and the norm of their Fourier transform outside a disk of radius $c$, where $c$ is the chosen bandlimit. We have compared our method with the FFBsPCA algorithm~\cite{zhao2014fast}, which is considered state-of-the-art for performing steerable-PCA on single-particle cryo-EM projection images, and have shown that our method is both faster and more accurate (for sufficiently small values of $T$). In addition, our method enjoys rigorous error bounds throughout its various steps, whereas in contrast, the FFBsPCA algorithm provides no analytic guarantees on its accuracy. We mention that classical operations of windowing and filtering can be subsumed in our procedure to ensure that the images fulfil the requirements of space-frequency localization.

As image resolutions get higher, investigating more efficient methods for processing image datasets is an important ongoing research task. As the running times of our algorithms are mostly dominated by the task of computing the PSWFs expansion coefficients, reducing the computational complexity of this step from $O(L^3)$ to $O(L^2\log L)$ for each image, resulting in the same asymptotic complexity as the two-dimensional FFT, will be a significant progress. Since the radial part of the PSWFs is evaluated to a prescribed accuracy using a finite series of special polynomials which admit a recurrence relation (see \cite{shkolnisky2007prolate}), and since the expansion coefficients in terms of these polynomials are obtained from the eigenvectors of a tridiagonal matrix, it is possible to employ the methods described in~\cite{tygert2010recurrence} to derive an $O(L^2\log L)$ algorithm for computing the PSWFs expansion coefficients of a function. In addition, as mentioned earlier, the NFFT, which was employed to map the Cartesian grid samples to a polar grid, is a major time-consuming component. In this context, we mention that gridding methods (see for example~\cite{rosenfeld1998optimal}), when applied carefully, may accelerate such a mapping, and thus reduce the overall running time of our algorithms.

\section*{Acknowledgements }
This research was supported by THE ISRAEL SCIENCE
FOUNDATION grant No. 578/14, and by Award Number R01GM090200 from the NIGMS.
\begin{appendices}
\section{PSWFs expansion error bound} \label{appendix:PSWFs expansion error bound}
Let us define the restriction of an image $I(x)$ to the unit disk by
\begin{equation}
\bar{I}(x) \triangleq
\begin{cases}
I(x) & x \in \mathbf{D}, \\
0 & x \notin \mathbf{D},
\end{cases}
\end{equation}
and denote the Fourier transform of $\bar{I}(x)$ by
\begin{equation}
\bar{I}^{\mathcal{F}}(\omega) \triangleq \mathcal{F}\left\lbrace \bar{I}\right\rbrace (\omega) = \int_{\mathbb{R}^2} \bar{I}(x) e^{\imath \omega x} dx .
\end{equation}
Since the Fourier transform is a linear operator, we can write
\begin{equation}
\bar{I}^{\mathcal{F}}(\omega) = \mathcal{F}\left\lbrace  I - \left(I-\bar{I}\right) \right\rbrace (\omega) = {I}^{\mathcal{F}}(\omega) + \mathcal{F}\left\lbrace I-\bar{I} \right\rbrace (\omega),
\end{equation}
where we defined the Fourier transform of $I(x)$ by ${I}^{\mathcal{F}}(\omega)$ .
Now, it is clear that
\begin{align}
\left\Vert \bar{I}^{\mathcal{F}}(\omega)\right\Vert_{\mathcal{L}^2(\omega \notin c\mathbf{D})} &\leq
\left\Vert {I}^{\mathcal{F}}(\omega)\right\Vert_{\mathcal{L}^2(\omega \notin c\mathbf{D})} + \left\Vert \mathcal{F}\left\lbrace I-\bar{I} \right\rbrace (\omega) \right\Vert_{\mathcal{L}^2(\omega \notin c\mathbf{D})} \\ &\leq
\left\Vert {I}^{\mathcal{F}}(\omega)\right\Vert_{\mathcal{L}^2(\omega \notin c\mathbf{D})} + \left\Vert \mathcal{F}\left\lbrace I-\bar{I} \right\rbrace (\omega) \right\Vert_{\mathcal{L}^2(\mathbb{R}^2)}.
\end{align}
Next, by employing Parseval's identity we obtain
\begin{equation}
\left\Vert \mathcal{F}\left\lbrace I-\bar{I} \right\rbrace (\omega) \right\Vert_{\mathcal{L}^2(\mathbb{R}^2)} = 2\pi \left\Vert I(x)-\bar{I}(x) \right\Vert_{\mathcal{L}^2(\mathbb{R}^2)},
\end{equation}
and thus, by using our space/frequency concentration assumptions on the image $I(x)$, we have that
\begin{equation}
\left\Vert \bar{I}^{\mathcal{F}}(\omega)\right\Vert_{\mathcal{L}^2(\omega \notin c\mathbf{D})} \leq \delta_c + 2\pi\varepsilon. \label{eq:trunc imag norm}
\end{equation}
Therefore, we conclude that $\bar{I}(x)$ is $(1,\bar{\varepsilon})$-concentrated in space, and its Fourier transform $\bar{I}^{\mathcal{F}}(\omega)$ is $(c,\bar{\delta}_c)$-concentrated, where
\begin{equation}
\bar{\varepsilon} = 0, \quad \bar{\delta_c} = \delta_c + 2\pi\varepsilon. \label{eq:trunc epsilon and delta}
\end{equation}
Now, it follows from~\cite{landa2016approximation} that the approximation error of $\bar{I}$ satisfies
\begin{equation}
{\left\Vert \bar{I}(x) - \underset{N,n \in \Omega_T}{\sum}\hat{a}_{N,n}{\psi}_{N,n}^c(x) \right\Vert}_{\mathcal{L}^2\left( \mathbf{D}\right)}
\leq \left( \bar{\varepsilon} + \frac{\bar{\delta_c}}{2\pi} \right) T + \eta \left(\frac{c}{2\pi L}\right)^2 \sqrt{\sum_{\frac{\vec{k}}{L}\notin\mathbf{D}}\left|\bar{I}(\frac{k}{L})\right|^{2}} +\frac{2}{\pi}\bar{\delta}_c, \label{eq:Total approx err}
\end{equation}
where $T$ is the truncation parameter defined in~\eqref{eq:PSWFs truncation rule}, and $\eta$ is a constant no bigger than ${2 \pi^2 L}/{c}$.
Finally, since $I_m(x)$ and $\bar{I}_m(x)$ coincide on $\mathbf{D}$, we get using~\eqref{eq:Total approx err}
\begin{equation}
{\left\Vert I_m(x) - \underset{N,n \in \Omega_T}{\sum}\hat{a}_{N,n}^m{\psi}_{N,n}^c(x) \right\Vert}_{\mathcal{L}^2\left( \mathbf{D}\right)}
\leq \left( \varepsilon + \frac{\delta_c}{2\pi} \right) \left( T + 4\right).
\end{equation}

\section{A bound for $\mathcal{N}_\theta^\ell$} \label{appendix:Number of quadrature nodes in theta}
Recall that we choose the number of angular nodes per radius, denoted $\mathcal{N}_\theta^\ell$, such that it satisfies
\begin{equation}
\sum_{|j| > \mathcal{N}_\theta^{\ell}} \left\vert J_j(2cr_\omega^\ell \rho) \right\vert < C_1 \vartheta_q, \label{eq:angular nodes condition}
\end{equation}
for every $\rho \in [0,1]$.
Using a bound for the Bessel functions~\cite{abramowitz1964handbook} together with the fact that $\left| \rho \right| \leq 1$, we get
\begin{equation}
\left\vert J_j(2cr_\omega^\ell \rho) \right\vert \leq
\left( \frac{2cr_\omega^\ell \rho}{2} \right)^j \frac{1}{\Gamma(j+1)} \leq
\left( {cr_\omega^\ell} \right)^j \frac{1}{\left( j+1 \right)!},
\end{equation}
and by using Stirling's approximation~\cite{abramowitz1964handbook} (for $n> 1$)
\begin{equation}
n! \geq \sqrt{2\pi n}\left( \frac{n}{e} \right)^n > e\left( \frac{n}{e} \right)^n
\end{equation}
we obtain (for $j>0$)
\begin{equation}\label{eq:bound of Jj}
\left\vert J_j(2cr_\omega^\ell \rho) \right\vert \leq
\frac{1}{c r_\omega^\ell e} \left( \frac{c r_\omega^\ell e}{j+1} \right)^{j+1}.
\end{equation}
Then, using the fact that the Bessel functions of the first kind satisfy
\begin{equation}
J_{-n}(x) = (-1)^n J_n(x),
\end{equation}
we can write
\begin{equation}
\sum_{|j| > \mathcal{N}_\theta^{\ell}} \left\vert J_j(2cr_\omega^\ell \rho) \right\vert =
2\sum_{j > \mathcal{N}_\theta^{\ell}} \left\vert J_j(2cr_\omega^\ell \rho) \right\vert \leq
\frac{2}{c r_\omega^\ell e} \sum_{j > \mathcal{N}_\theta^{\ell}} \left( \frac{c r_\omega^\ell e}{j+1} \right)^{j+1}. \label{eq:bound for sum of Jj}
\end{equation}
We define
\begin{equation}
\gamma_\ell \triangleq {c r_\omega^\ell e}
\end{equation}
and choose the number of quadrature nodes per radius as
\begin{equation}
\mathcal{N}_\theta^{\ell} = \lceil \gamma_\ell \rceil + d,
\end{equation}
where $d$ is some positive integer (to be determined shortly) and $\lceil \cdot \rceil$ denotes the rounding up operation.
It can be easily verified that
\begin{equation}\label{eq:auxgamma}
\frac{\gamma_\ell}{j+1} \leq \frac{\gamma_\ell}{\gamma_\ell + d}
\end{equation}
whenever $j>\mathcal{N}_\theta^{\ell}$.
Using~\eqref{eq:bound for sum of Jj} and~\eqref{eq:auxgamma} we get that
\begin{equation}
\sum_{|j| > \mathcal{N}_\theta^{\ell}} \left\vert J_j(2cr_\omega^\ell \rho) \right\vert \leq
\frac{2}{\gamma_\ell} \sum_{j > \mathcal{N}_\theta^{\ell}} \left( \frac{\gamma_\ell}{\gamma_\ell + d} \right)^{j+1} \leq 2 \left( 1 + \frac{d}{\gamma_\ell}\right)^{-\mathcal{N}_\theta^{\ell}} \leq
2 \left( 1 + \frac{d}{\gamma_\ell}\right)^{-\gamma_\ell-d} \leq 2e^{-d}
\end{equation}
where we have used the inequality
\begin{equation}
\left( 1 + \frac{a}{b}\right)^{-b} < e^{-\frac{ab}{a+b}},
\end{equation}
from \cite{abramowitz1964handbook}, with $a = (\gamma_\ell+d)\frac{d}{\gamma_\ell}$ and $b = \gamma_\ell+d$.
Finally, we can see that in order to satisfy \eqref{eq:angular nodes condition} it is sufficient to choose
\begin{equation}
d = d(\vartheta_q) \geq \log{\vartheta_q^{-1}} + \log{\frac{2}{C_1}},
\end{equation}
and thus
\begin{equation}
\mathcal{N}_\theta^{\ell} \geq {c r_\omega^\ell e} + \log{\vartheta_q^{-1}} + \log{\frac{2}{C_1}} + 1.
\end{equation}

\section{Behavior and decay properties of $\lambda^{2c}_{0,k}$} \label{appendix:Behevaior of lambda}
We start by reviewing some known results on the behavior of the eigenvalues of the PSWFs in the one-dimensional setting, where they have been thoroughly investigated (see~\cite{osipov2013prolate} and references therein). The most well-known characterization of these eigenvalues is that they can be divided into three distinct regions of behavior (as a function of their index $n$) - a flat region, where the (normalized) eigenvalues are essentially $1$, a transitional region, where they decay from values close to $1$ to values close to $0$, and a super-exponential decay region, where they are very close to $0$ and decay as $\sim e^{-n\log{n}}$.
In addition, it is known that if we choose all eigenvalues that are greater than some small $\epsilon$, then there are about ${2c}/{\pi}$ eigenvalues from the flat region, $O(\log{\left(\frac{1-\epsilon}{\epsilon}\right)}\log{c})$ eigenvalues from the transitional region, and $o(\log{c})$ eigenvalues from the decay region (see \cite{landau1980eigenvalue} for a precise  formulation). Thus, the number of eigenvalues greater then $\epsilon$ is dominated by the number of eigenvalues in the flat region, which is ${2c}/{\pi}$. As for the eigenvalues of the two-dimensional PSWFs, results in $\cite{serkh2015prolates}$ indicate that as in the one-dimensional setting, the eigenvalues can be similarly divided into three distinct regions - flat, transitional, and super-exponential decay regions. Correspondingly, the number of significant eigenvalues is dominated by the number of eigenvalues in the flat region. Since $\lambda_{0,k}^{2c} \propto \alpha^{2c}_{0,k}$ (see~\eqref{eq:PSWFs truncation rule}), it is clear that in order to satisfy condition~\eqref{eq:numerical integration radial nodes condition} we need to determine the number of terms in the flat region of $\left| \lambda_{0,k}^{2c} \right|$. To this end, we follow \cite{landau1975szego}, which provides similar results for general non-hermitian Toeplitz integral operators (see also \cite{landau1980eigenvalue} and \cite{xiao2001prolate}), and consider the sum
\begin{equation}
\sum_{k=0}^{\infty} \left|\lambda_{0,k}^{2c}\right|^2,
\end{equation}
which is approximately equal to the number of values of $\left|\lambda_{0,k}^{2c}\right|$ which are close to $1$ (denoted as the flat region). For simplicity of the presentation, we evaluate this sum for a bandlimit of $c$, and eventually, replace $c$ with $2c$.
From~\cite{slepian1964prolate}, the radial functions $R_{N,n}^c(r)$ in \eqref{eq:PSWFs complex form} are real-valued, and are obtained as the solutions to the integral equation
\begin{equation}
\beta R(r)=\int_0^1 R(\rho) J_N(cr\rho) \rho d\rho ,\quad r\in [0,1], \label{eq:PSWFs radial part integral eq}
\end{equation}
where $J_N(x)$ is the Bessel function of the first kind of order $N$. The eigenvalues $\alpha_{N,n}^c$ and $\beta_{N,n}^c$ of \eqref{eq:Basic PSWF eq} and \eqref{eq:PSWFs radial part integral eq} are related by
\begin{equation}
\alpha_{N,n}^c = 2\pi\imath^N\beta_{N,n}^c.
\end{equation}
By substituting $\varphi(r)=R(r)\sqrt{r}$ and $\gamma=\beta\sqrt{c}$ into~\eqref{eq:PSWFs radial part integral eq}, we obtain the integral equation
\begin{equation}
\gamma \varphi(r)=\int_0^1 \varphi(\rho) J_N(cr\rho) \sqrt{cr\rho} d\rho ,\quad r\in [0,1], \label{eq:phi radial equation}
\end{equation}
whose eigenvalues and eigenfunctions are $\gamma_{N,n}^c$ and $\varphi_{N,n}^c(r)$, respectively. Equation~\eqref{eq:phi radial equation} was analysed in~\cite{slepian1964prolate}, where it is established that the eigenfunctions $\varphi_{N,n}^c(r)$ constitute a complete orthonormal system in $\mathcal{L}^2\left[ 0,1\right]$. Therefore, it follows that we have the identity
\begin{equation}
J_N(cr\rho) \sqrt{cr\rho} = \sum_{k=1}^{\infty} \gamma_{N,n}^c \varphi_{N,n}^c(r) \varphi_{N,n}^c(\rho),
\end{equation}
for $r$ and $\rho$ both in $[0,1]$. If we notice that $\lambda_{N,n}^c = \sqrt{c}\gamma_{N,n}^c$, and take $N=0$, we have
\begin{equation}
J_0(cr\rho) c\sqrt{r\rho} = \sum_{k=1}^{\infty} \lambda_{0,k}^c \varphi_{0,k}^c(r) \varphi_{0,k}^c(\rho).
\end{equation}
Next, we take the squared absolute value of both sides of the equation above, followed by double integration (in $r$ and $\rho$) to obtain
\begin{equation}
\int_{0}^{1} \int_{0}^{1} J_0^2(cr\rho) c^2 r\rho \; dr d\rho = \sum_{k=0}^{\infty} \left|\lambda_{0,k}^c\right|^2. \label{eq:J_0^2 double integral}
\end{equation}
By evaluating the left hand-side of~\eqref{eq:J_0^2 double integral} using known integral identities of the Bessel functions~\cite{abramowitz1964handbook}, and after some manipulation, one can verify that
\begin{equation}
\sum_{k=0}^{\infty} \left|\lambda_{0,k}^c\right|^2 = \frac{c^2}{4} \left( J_0^2(c) - J_2(c)J_0(c) + 2J_1^2(c) \right).
\end{equation}
Now, using the asymptotic approximation (see~\cite{abramowitz1964handbook})
\begin{equation}
J_m(x) \sim \sqrt{\frac{2}{\pi x}} \cos(x - \frac{m \pi}{2} - \frac{\pi}{4}),
\end{equation}
valid for $ x \gg \left| m^2 - 1/4 \right| $, we can write
\begin{align}
\sum_{k=0}^{\infty} \left|\lambda_{0,k}^c\right|^2 &\sim
\frac{c}{2\pi} \left( \cos^2(c-\frac{\pi}{4}) - \cos(c-\pi-\frac{\pi}{4})\cos(c-\frac{\pi}{4}) + 2\cos^2(c-\frac{\pi}{2}-\frac{\pi}{4}) \right) \nonumber \\
&= \frac{c}{2\pi} \left( 2\cos^2(c-\frac{\pi}{4}) + 2\sin^2(c-\frac{\pi}{4}) \right) \nonumber \\
&= \frac{c}{\pi}. \label{eq:sum lambda final}
\end{align}
Therefore, the expression in~\eqref{eq:sum lambda final} establishes that for large values of $c$, the number of terms in the set $\left\lbrace \left|\lambda_{0,n}^{2c}\right| \right\rbrace$ which are close to $1$ is about ${2c}/{\pi}$ (see also Figure~\ref{fig:lambda_0_n spectrum}).

\end{appendices}

\bibliographystyle{plain}
\bibliography{mybib}

\begin{thebibliography}{10}

\bibitem{aspire}
Algorithms for single particle reconstruction.
\newblock \url{http://spr.math.princeton.edu/}.

\bibitem{abramowitz1964handbook}
Milton Abramowitz and Irene~A Stegun.
\newblock {\em Handbook of mathematical functions: with formulas, graphs, and
  mathematical tables}.
\newblock Courier Corporation, 1964.

\bibitem{dutt1993fast}
Alok Dutt and Vladimir Rokhlin.
\newblock Fast {Fourier} transforms for nonequispaced data.
\newblock {\em SIAM Journal on Scientific computing}, 14(6):1368--1393, 1993.

\bibitem{ferraro1988relationship}
Mario Ferraro and Terry~M Caelli.
\newblock Relationship between integral transform invariances and {Lie} group
  theory.
\newblock {\em JOSA A}, 5(5):738--742, 1988.

\bibitem{fessler2003nonuniform}
Jeffrey~A Fessler and Bradley~P Sutton.
\newblock Nonuniform fast {Fourier} transforms using min-max interpolation.
\newblock {\em Signal Processing, IEEE Transactions on}, 51(2):560--574, 2003.

\bibitem{Frank}
Joachim Frank.
\newblock {\em Three-Dimensional Electron Microscopy of Macromolecular
  Assemblies: Visualization of Biological Molecules in Their Native State}.
\newblock Oxford, 2006.

\bibitem{freeman1991design}
William~T. Freeman and Edward~H Adelson.
\newblock The design and use of steerable filters.
\newblock {\em IEEE Transactions on Pattern Analysis \& Machine Intelligence},
  13(9):891--906, 1991.

\bibitem{greengard2004accelerating}
Leslie Greengard and June-Yub Lee.
\newblock Accelerating the nonuniform fast {Fourier} transform.
\newblock {\em SIAM review}, 46(3):443--454, 2004.

\bibitem{hilai1994recognition}
Ran Hilai and Jacob Rubinstein.
\newblock Recognition of rotated images by invariant {Karhunen}--{Lo{\'e}ve}
  expansion.
\newblock {\em JOSA A}, 11(5):1610--1618, 1994.

\bibitem{jogan2002karhunen}
Matjaz Jogan, Emil Zagar, and Aleˇs Leonardis.
\newblock {Karhunen--Loeve} expansion of a set of rotated templates.
\newblock {\em IEEE transactions on image processing: a publication of the IEEE
  Signal Processing Society}, 12(7):817--825, 2002.

\bibitem{kanatani2012group}
Kenichi Kanatani.
\newblock {\em Group-theoretical methods in image understanding}, volume~20.
\newblock Springer Science \& Business Media, 2012.

\bibitem{keiner2009using}
Jens Keiner, Stefan Kunis, and Daniel Potts.
\newblock Using {NFFT} 3 -- a software library for various nonequispaced fast
  {Fourier} transforms.
\newblock {\em ACM Transactions on Mathematical Software (TOMS)}, 36(4):19,
  2009.

\bibitem{landa2016approximation}
Boris Landa and Yoel Shkolnisky.
\newblock Approximation scheme for essentially bandlimited and
  space-concentrated functions on a disk.
\newblock {\em Applied and Computational Harmonic Analysis}, 2016.

\bibitem{landau1967necessary}
Henry~J. Landau.
\newblock Necessary density conditions for sampling and interpolation of
  certain entire functions.
\newblock {\em Acta Mathematica}, 117(1):37--52, 1967.

\bibitem{landau1975szego}
Henry~J. Landau.
\newblock On {Szeg{\"o}'s} eingenvalue distribution theorem and non-hermitian
  kernels.
\newblock {\em Journal d'Analyse Math{\'e}matique}, 28(1):335--357, 1975.

\bibitem{landau1961prolate}
Henry~J Landau and Henry~O Pollak.
\newblock Prolate spheroidal wave functions, {Fourier} analysis and uncertainty
  —- {II}.
\newblock {\em Bell System Technical Journal}, 40(1):65--84, 1961.

\bibitem{landau1962prolate}
Henry~J Landau and Henry~O Pollak.
\newblock Prolate spheroidal wave functions, {Fourier} analysis and uncertainty
  —- {III}: The dimension of the space of essentially time-and band-limited
  signals.
\newblock {\em Bell System Technical Journal}, 41(4):1295--1336, 1962.

\bibitem{landau1980eigenvalue}
Henry~J. Landau and Harold Widom.
\newblock Eigenvalue distribution of time and frequency limiting.
\newblock {\em Journal of Mathematical Analysis and Applications},
  77(2):469--481, 1980.

\bibitem{Lawson04012016}
Catherine~L. Lawson, Ardan Patwardhan, Matthew~L. Baker, Corey Hryc,
  Eduardo~Sanz Garcia, Brian~P. Hudson, Ingvar Lagerstedt, Steven~J. Ludtke,
  Grigore Pintilie, Raul Sala, John~D. Westbrook, Helen~M. Berman, Gerard~J.
  Kleywegt, and Wah Chiu.
\newblock {EMDataBank} unified data resource for {3DEM}.
\newblock {\em Nucleic Acids Research}, 44(D1):D396--D403, 2016.

\bibitem{lenz1989group}
Reiner Lenz.
\newblock Group-theoretical model of feature extraction.
\newblock {\em JOSA A}, 6(6):827--834, 1989.

\bibitem{lenz1990group}
Reiner Lenz.
\newblock Group invariant pattern recognition.
\newblock {\em Pattern Recognition}, 23(1):199--217, 1990.

\bibitem{lenz1990group2}
Reiner Lenz.
\newblock {\em Group theoretical methods in image processing}.
\newblock Springer-Verlag New York, Inc., 1990.

\bibitem{mallick2005ace}
Satya~P. Mallick, Bridget Carragher, Clinton~S. Potter, and David~J. Kriegman.
\newblock {ACE:} automated {CTF} estimation.
\newblock {\em Ultramicroscopy}, 104(1):8--29, 2005.

\bibitem{osipov2013prolate}
Andrei Osipov, Vladimir Rokhlin, and Hong Xiao.
\newblock Prolate spheroidal wave functions of order zero.
\newblock {\em Springer Ser. Appl. Math. Sci}, 187, 2013.

\bibitem{perona1995deformable}
Pietro Perona.
\newblock Deformable kernels for early vision.
\newblock {\em Pattern Analysis and Machine Intelligence, IEEE Transactions
  on}, 17(5):488--499, 1995.

\bibitem{petersen1962sampling}
Daniel~P Petersen and David Middleton.
\newblock Sampling and reconstruction of wave-number-limited functions in
  $n$-dimensional euclidean spaces.
\newblock {\em Information and control}, 5(4):279--323, 1962.

\bibitem{ponce2011computing}
Colin Ponce and Amit Singer.
\newblock Computing steerable principal components of a large set of images and
  their rotations.
\newblock {\em Image Processing, IEEE Transactions on}, 20(11):3051--3062,
  2011.

\bibitem{potts2001fast}
Daniel Potts, Gabriele Steidl, and Manfred Tasche.
\newblock Fast {Fourier} transforms for nonequispaced data: A tutorial.
\newblock In {\em Modern sampling theory}, pages 247--270. Springer, 2001.

\bibitem{rosenfeld1998optimal}
Daniel Rosenfeld.
\newblock An optimal and efficient new gridding algorithm using singular value
  decomposition.
\newblock {\em Magnetic Resonance in Medicine}, 40(1):14--23, 1998.

\bibitem{serkh2015prolates}
Kirill Serkh.
\newblock On generalized prolate spheroidal functions.
\newblock Technical Report TR-1519, Department of Mathematics, Yale University,
  2015.

\bibitem{shkolnisky2007prolate}
Yoel Shkolnisky.
\newblock Prolate spheroidal wave functions on a disc —- integration and
  approximation of two-dimensional bandlimited functions.
\newblock {\em Applied and Computational Harmonic Analysis}, 22(2):235--256,
  2007.

\bibitem{slepian1964prolate}
David Slepian.
\newblock Prolate spheroidal wave functions, {Fourier} analysis and uncertainty
  -— {IV}: extensions to many dimensions; generalized prolate spheroidal
  functions.
\newblock {\em Bell System Technical Journal}, 43(6):3009--3057, 1964.

\bibitem{slepian1961prolate}
David Slepian and Henry~O Pollak.
\newblock Prolate spheroidal wave functions, {Fourier} analysis and uncertainty
  —- {I}.
\newblock {\em Bell System Technical Journal}, 40(1):43--63, 1961.

\bibitem{tygert2010recurrence}
Mark Tygert.
\newblock Recurrence relations and fast algorithms.
\newblock {\em Applied and Computational Harmonic Analysis}, 28(1):121--128,
  2010.

\bibitem{vonesch2015steerable}
C{\'e}dric Vonesch, Fr{\'e}d{\'e}ric Stauber, and Michael Unser.
\newblock Steerable {PCA} for rotation-invariant image recognition.
\newblock {\em SIAM Journal on Imaging Sciences}, 8(3):1857--1873, 2015.

\bibitem{xiao2001prolate}
Hong Xiao, Vladimir Rokhlin, and Norman Yarvin.
\newblock Prolate spheroidal wavefunctions, quadrature and interpolation.
\newblock {\em Inverse problems}, 17(4):805, 2001.

\bibitem{zhao2014fast}
Zhizhen Zhao, Yoel Shkolnisky, and Amit Singer.
\newblock Fast steerable principal component analysis.
\newblock {\em IEEE Transactions on Computational Imaging}, 2(1):1--12, 2016.

\bibitem{zhao2013fourier}
Zhizhen Zhao and Amit Singer.
\newblock {Fourier--Bessel} rotational invariant eigenimages.
\newblock {\em JOSA A}, 30(5):871--877, 2013.

\end{thebibliography}

\end{document}